\documentclass{article} % For LaTeX2e

\usepackage{iclr2026_arxiv,times}
\usepackage{url}
\usepackage{microtype}
\usepackage{xspace}
\usepackage{xcolor}

\usepackage{amsmath}
\usepackage{amssymb}
\usepackage{amsfonts}
\usepackage{amsthm}
\usepackage{bm}
\usepackage{nicefrac}

\usepackage{graphicx}
\usepackage{subcaption}
\usepackage{wrapfig}
\usepackage{tikz}

\usepackage{booktabs}
\usepackage{colortbl}
\usepackage{multirow}
\usepackage{arydshln}
\usepackage{tabularx}

\usepackage{algorithm}
\usepackage{algpseudocode}

\usepackage{enumitem}
\usepackage{float}
\usepackage{pifont}
\usepackage{tcolorbox}
\usepackage{fontawesome5} 
\usepackage{hyperref}

\definecolor{phaseblue}{RGB}{31,78,121}
\definecolor{phasegreen}{RGB}{34,139,34}
\definecolor{phaseorange}{RGB}{230,126,34}

\definecolor{keycolor}{RGB}{0,51,179}       % dark blue for keys
\definecolor{tagcolor}{RGB}{179,0,0}        % dark red for tags
\definecolor{varcolor}{RGB}{0,100,0}        % dark green for variables

\definecolor{mygray}{gray}{.9}
\definecolor{ggray}{RGB}{127,127,127}
\definecolor{reda}{RGB}{192,0,0}
\definecolor{redb}{RGB}{217,148,143}
\definecolor{myyellow}{RGB}{190,144,0}
\definecolor{mygreen}{RGB}{80,100,40}
\definecolor{myblue}{RGB}{30,90,100}
\definecolor{tabhighlight}{HTML}{e5e5e5}

\definecolor{sectioncolor}{RGB}{0,51,153}
\definecolor{harmcolor}{RGB}{200,0,0}
\definecolor{chatgray}{RGB}{80,80,80}

 \hypersetup{
 	colorlinks   = true,
 	urlcolor     = blue,
 	linkcolor    = reda,
 	citecolor   = blue
 }
\newcommand{\ourapproach}{PivoARL\xspace}

\newtheorem{proposition}{Proposition}

\title{Agent Reinforcement Learning via Pivotal-Aware Self-Feedback Retry}

\iclrfinalcopy

\author{Weiyang Guo$^1$\, Zesheng Shi$^1$\, Longhui Zhang$^1$\,  Zeen Zhu$^{1}$\,  Min Zhang$^1$\, Jing Li$^1$\textsuperscript{\faEnvelope[regular]}\\
  $^1$Harbin Institute of Technology, Shenzhen, China\\
  \texttt{weiyangguo2024@gmail.com} \quad \texttt{jingli.phd@hotmail.com}  \\
}

\begin{document}

\maketitle

\begin{abstract}
Large language model (LLM) agents have shown strong decision-making capabilities in long-horizon interactive tasks, yet they still struggle to effectively leverage failed trajectories: full retries incur high interaction costs, while experience retrieval tends to dilute critical experience signals. To address this, we propose \ourapproach, a self-feedback retry framework for experience exploitation in LLM agents. \ourapproach identifies the pivotal erroneous turn through structured reflection and performs local retry only from the corresponding pivotal state, thereby reusing the correct prefix and reducing redundant interactions.  From an information-gain perspective, we further show that pivotal retry concentrates useful experience signals near the error boundary, mitigating the signal dilution caused by state-agnostic experience utilization. Based on this insight, we design a pivotal-aware credit assignment mechanism that rewards correct prefixes while isolating erroneous suffixes, and optimize reflection quality through implicit reflection returns. We conduct a systematic evaluation on 4 agent tasks and 7 search-based QA benchmarks. Results show that \ourapproach achieves significant improvements on Pass@2/3 across all tasks, with an average gain of about 11.5\% over MetaRL. Moreover, benefiting from contrastive preference signals induced by pivotal turns, \ourapproach also consistently improves Pass@1 on over 80\% of the tasks. On Minesweeper environment, \ourapproach improves over GiGPO by more than 45\% and reduces interaction turns by about 42\% on average compared with full-retry methods. Code is available at \url{https://github.com/yuki-younai/PivoARL}.
\end{abstract}

\section{Introduction}
\label{sec:introduction}

Large language models (LLMs) have evolved from static question-answering systems into versatile \emph{agents} that perceive, reason, and act across multi-turn interactions with external environments~\citep{yao2023react,shinn2023reflexion,guo2025mtsa}.
These agents navigate simulated households~\citep{shridhar2020alfworld}, plan multi-step web browsing sessions~\citep{yao2022webshop}, code development
 ~\citep{guo2026e3tirenhancedexperienceexploitation}, and solve complex tasks that demand long-horizon planning and adaptive decision-making. However, training efficiency remains a bottleneck due to high interaction costs and sparse rewards~\citep{landscape_arl}. A fundamental challenge persists: \textit{how to effectively mine experiences from past trajectories to facilitate robust learning?}

Despite its potential in long-horizon agent tasks~\citep{wang2025ragen}, experience-based reinforcement learning (RL) faces dual challenges~\citep{wang2026ragen2reasoningcollapseagentic}. First, there is a trade-off between precision and efficiency in experience utilization: static library-based methods (e.g., SkillsRL~\citep{xia2026skillrlevolvingagentsrecursive}) struggle to adapt to dynamic task evolution, while dynamic paradigms are often susceptible to interference when processing large-scale experiences. Second, credit assignment remains a major challenge in long-horizon agent tasks. Sparse final feedback from failed trajectories makes it difficult to localize the exact erroneous step~\citep{lamer}, causing negative advantages to be mistakenly assigned to correct prefix decisions. This weakens the learning of useful experiences and may lead to training instability or even capability degradation~\citep{li2026mol,erl}.

% Second, experience internalization remains limited. Although self-feedback improves success rates over multiple trials~\citep{lamer}, it often fails to enhance zero-shot performance. Existing approaches typically rely on SFT~\citep{li2026mol,erl} or offline RL~\citep{empo2, critiquegrpo} to inject experiences directly into model parameters. However, due to distribution shift, such methods frequently lead to training instability or even model degradation.

\begin{figure*}[t]
    \centering 
    \includegraphics[width=0.98\linewidth]
    {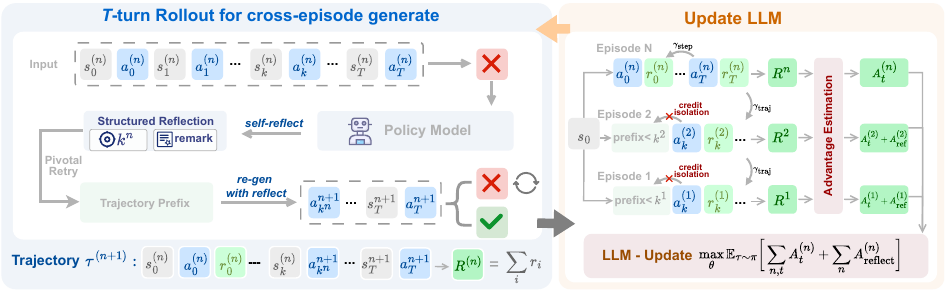}
    \caption{\textbf{The Pivotal-Aware Self-Feedback Retry (\ourapproach) framework}. The agent identifies pivotal errors via structured self-reflection, retries from the pivotal state with the correct prefix reused, and updates the LLM through pivotal-isolated cross-episode credit assignment.}
    \label{framwork}
\end{figure*}

% === Our approach ===
In this study, we propose \ourapproach, a general reinforcement learning paradigm for agents. This paradigm identifies the pivotal turn responsible for task failure through a structured reflection mechanism and initiates targeted local retries from that specific state, thereby maximizing the reuse of correct historical prefixes. To optimize the learning process, we design a cross-episode prefix credit assignment mechanism and implicit reflection returns. These components enable the agent to deeply mine effective signals from failed trajectories and ultimately transform them into reflection-guided, high-quality successful experiences.  From an information-gain perspective, we further demonstrate that pivotal retries concentrate useful experience signals near the error boundary, effectively mitigating the signal dilution commonly found in global reflections.

Experimental results show that \ourapproach brings consistent gains across four agent environments and seven search-based QA benchmarks. On agent tasks, \ourapproach improves over GiGPO by about 10.3\%, 15.3\%, and 14.7\% on Pass@1/2/3, respectively; on search-based QA tasks, it improves over GRPO by about 20.5\%. Meanwhile, benefiting from the efficient reuse mechanism of pivotal retry, \ourapproach reduces interaction cost by about 44\% compared with MetaRL. We further conduct detailed ablation studies to quantify the contribution of each component, and perform additional analyses to examine the test-time and training-time scaling behavior of \ourapproach.
% === Contributions ===

Our contributions are as follows:
\begin{itemize}[leftmargin=*,itemsep=2pt,topsep=3pt]
    \item \textbf{Pivotal-Aware Retry.} We propose a self-feedback retry mechanism that identifies the earliest erroneous turn in a failed trajectory and restarts from this pivotal state, reusing the correct prefix to reduce redundant interactions and improve exploitation.
    \item \textbf{Information Gain Theory.} We analyze experience learning from an information-gain perspective, showing that global reflections or retrieved experiences are easily diluted in long trajectories, while pivotal retry concentrates useful guidance around the critical error boundary.
    \item \textbf{State-of-the-Art Performance.} \ourapproach improves Pass@k across four agent environments and seven search-based QA benchmarks, achieving about 16\% average relative gain over default RL baselines, while demonstrating stronger test-time and training-time scaling.
\end{itemize}

\section{Preliminary}
\label{sec:preliminary}

\subsection{Notation and problem setup}
\label{sec:notation}

We formulate LLM agent interaction as a Markov decision process~\citep{lamer} $\mathcal{M} = (\mathcal{S}, \mathcal{A}, P, R)$.
Given a task $x \sim D$, the agent operates over $T$ steps: at step $t$, it observes  state $\bm{s}_t \in \mathcal{S}$, generates action $\bm{a}_t \in \mathcal{V}^{n}$ via policy $\pi_\theta(\bm{a}_t \mid \bm{s}_t)$. The environment then provides a scalar reward $r_t \in \mathbb{R} $, observation $o_t$ and transitions to $\bm{s}_{t+1}$.
For each trajectory $\bm{\tau} = \{(\bm{s}_0, \bm{a}_0, \bm{r}_0), \ldots, (\bm{s}_{T-1}, \bm{a}_{T-1}, \bm{r}_{T-1})\}$, we utilize the total return $R(\tau)=\sum_{t=0}^{T-1} r_t$ as a episode-level reward.

\paragraph{Self feedback with experience cycle.}
Beyond scalar rewards, the historical trajectory $\bm{\tau}_0$ provides rich  feedback at each step (e.g., error messages, state descriptions), revealing \emph{why} the agent succeeded or failed~\citep{mrsearch}. Given the interaction context $\bm{c}=\sum_{t}^{T-1} \{(a_t, o_t)\}$ from an initial trajectory, the agent constructs an augmented context to extract the past experience information:
\begin{equation}
    \textbf{remark} = \pi_{\theta}\big(\cdot   |x,  \; \{\bm{c}\}\big)
    \label{eq:augmented_context}
\end{equation}
Where $\textbf{remark}$ represents the experience gained from the previous attempt, utilized in the next attempt~\citep{r3l} as a new enhanced task prompt $\bm{x}^+$. This enables the model to leverage past experience for optimized subsequent attempts.

\subsection{Agentic Reinforcement Learning}
\label{sec:group_rl}

Agentic Reinforcement Learning (ARL) typically adopts policy-gradient methods~\citep{schulman2017ppo} to optimize the agent policy $\pi_{\theta}$. We formulate the agentic RL training objective as: 
\begin{equation}
\max_{\pi_\theta} \mathbb{E}_{x \sim \mathcal{D}, \mathcal{\tau}  \sim \pi_\theta(\cdot|x)} \left[ \sum_{t=0}^{T-1} A_t \right] - \beta \mathbb{D}_{\text{KL}} \left[ \pi_\theta(\cdot) \| \pi_{\text{ref}}(\cdot) \right]
\end{equation}
where $\pi_{\text{ref}}$ is the reference LLM, and $\mathbb{D}_{\text{KL}}$ and $\beta$ denote the KL divergence and its coefficient, respectively. $A_t$ is the step-wise advantage, typically computed from both step-level and episode-level rewards within the group.

% Recent RL methods for LLMs adopt group-based training~\citep{grpo,dapo}.
% For task $x$, the agent samples $N$ trajectories $\{\bm{\tau}_i\} \sim \pi_{\theta_{\text{old}}}$, each receiving reward $R(\bm{\tau}_i)$.
% The advantage is estimated via group normalization:
% \begin{equation}
%     \hat{A}(\bm{\tau}_i) = \frac{R(\bm{\tau}_i) - \text{mean}\big(\{R(\bm{\tau}_j)\}_{j=1}^{N}\big)}{F_{\text{norm}}\big(\{R(\bm{\tau}_j)\}_{j=1}^{N}\big)},
%     \label{eq:grpo_advantage}
% \end{equation}
% and the policy is updated via a clipped surrogate objective:
% \begin{equation}
%     \mathcal{J}_{\text{GRPO}}(\theta) = \mathbb{E}_{\substack{x \sim p(\mathcal{X}) \\ \{\bm{\tau}_i\} \sim \pi_{\theta_\text{old}}}} \left[ \frac{1}{NT} \sum_{i=1}^{N} \sum_{t=1}^{T} \min\Big(\rho_t^{(i)} \hat{A}(\bm{\tau}_i),\; \text{clip}\big(\rho_t^{(i)}, 1\pm\epsilon\big) \hat{A}(\bm{\tau}_i)\Big) \right],
%     \label{eq:grpo}
% \end{equation}
% where $\rho_t^{(i)} = \pi_\theta(\bm{a}_t^{(i)} \mid \bm{s}_t^{(i)}, x) / \pi_{\theta_\text{old}}(\bm{a}_t^{(i)} \mid \bm{s}_t^{(i)}, x)$ is the importance sampling ratio.
% A key limitation is that all tokens within a trajectory share the same advantage $\hat{A}(\bm{\tau}_i)$, which cannot distinguish pivotal actions from irrelevant ones in multi-turn episodes spanning thousands of tokens.
% This motivates the dense credit assignment approach developed in~\S\ref{sec:method}.

% TODO: Uncomment when ready
\section{Method}
\label{sec:method}

\subsection{Overview: The \ourapproach Framework}
\label{sec:overview}

\begin{wrapfigure}[10]{r}{7cm}
\centering
\vspace{-8mm}
\includegraphics[width=0.5\textwidth]{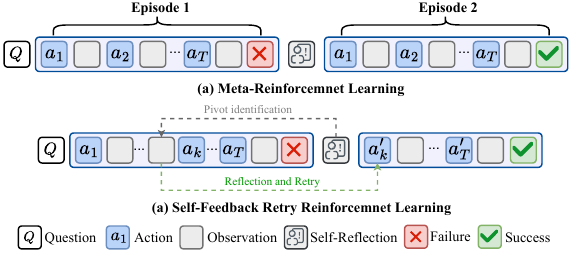}
\vspace{-3mm}
\caption{\label{fig2:self_feedback} Comparison between MetaRL and Pivotal-Aware Self-Feedback Retry RL. }
\end{wrapfigure}

The previous cross-episode paradigm~\citep{lamer} treats each retry as a full restart from $s_0$, discarding the correct prefix of the failed trajectory. We propose \textbf{Pivotal-Aware Self-Feedback}, which reuses the correct prefix and restarts from the pivotal error.

\paragraph{Pivotal Retry Training Framework.} In the training of \ourapproach, each trial consists of $N$ episodes sequentially generated by the agent:

\begin{equation}
\mathcal{T} = (\tau^{(0)}, \tau^{(1)}, \dots, \tau^{(N-1)}),  \ n \in [0, N-1]
\end{equation}

The rollout process terminates at $n$ if $\tau^n$ is successful. Otherwise, the agent initiates a new episode $\tau^{(n+1)}$.  Unlike prior work relying on past experiences for a full restart, our agent reflects to identify pivotal decision points, enabling partial trajectory reuse.

\paragraph{Structured Reflection and Pivotal Retry.}
After a failed episode $\tau^{(n)}$ of $T^{(n)}$ turns, the agent generates a structured reflection containing a textual remark and a \emph{pivotal turn index} $k^{(n)} \in [0, T^{(n)}-1]$ that identifies the earliest erroneous turn. The remark provides experience guidance injected into the original state during retry: 
\begin{equation}
    \Delta^{(n)}_{\text{reflect}} \sim \pi_\theta\bigl(\cdot \mid x, \bm{c}^n\bigr) = \bigl(k^{(n)}; \text{remark}^{(n)}\ \bigr).
    \label{eq:structured_reflection}
\end{equation}

Given the pivotal turn $k^{(n)}$, the next episode $\tau^{(n+1)}$ reuses the correct prefix of $\tau^{(n)}$, regenerating actions only from $k^{(n)}$:   

\begin{equation}
\begin{aligned}
\tau^{(n+1)} = \bigl(
&\underbrace{s_0^{(n)}, a_0^{(n)}, \ldots, s_{k^*}^{(n)}}_{\text{reused prefix}}, 
\underbrace{a_{k^*}^{(n+1)}, s_{k^*+1}^{(n+1)}, \ldots, a_{T^{(n+1)}-1}^{(n+1)}}_{\text{regenerated from } k^* (\text{where} \; k^* = k^{(n)}) }
\bigr).
\end{aligned}
\end{equation}

Where $T^{(n+1)}$ is the length of the new episode. This reduces the generation overhead for each retry from $T^{(n+1)}$ to $T^{(n+1)} - k^{(n)}$ tokens, while preserving the correct decision history as context.

\subsection{Pivotal-Aware Credit Assignment}
\label{sec:pivotal_retry}

\begin{figure*}[t]
    \centering 
    \includegraphics[width=0.98\linewidth]
    {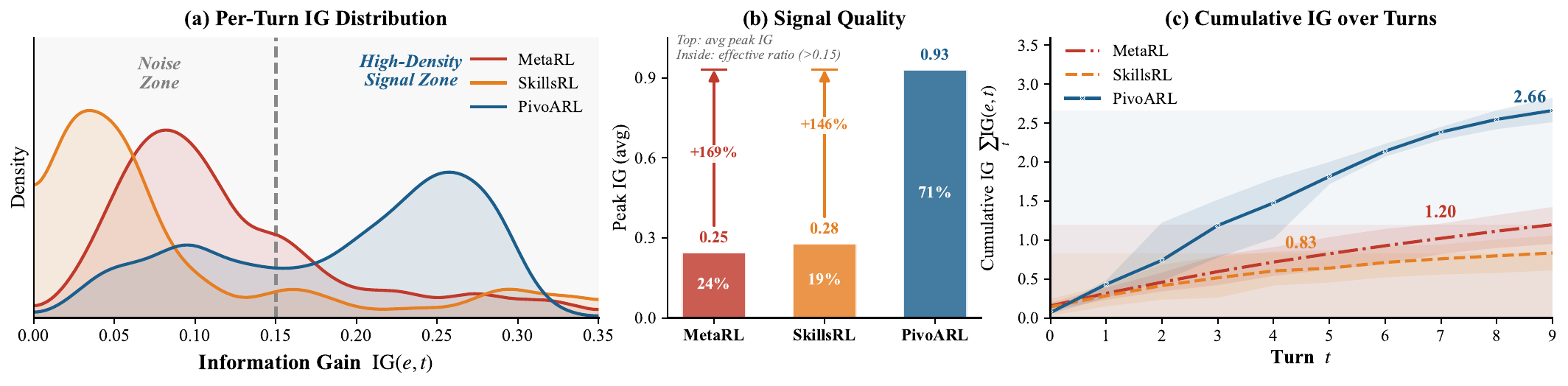}
    \caption{Information gain (IG) analysis of experience guidance. (a) \textbf{\textit{Per-turn IG density}}: our method concentrates IG in high-signal turns. (b) \textbf{\textit{Signal Quality}}: Peak IG and percentage of high-IG turns. (c) \textbf{\textit{Cumulative IG over turns}}: showing a characteristic leap at turn k for our method.}
    \label{framwork}
\end{figure*}

\paragraph{Credit Assignment with Pivotal Isolation.} To propagate credit both within and across episodes, we decompose credit return $G_{t}^{(n)}$ into intra-episode returns $g_{t}^{(n)}$ and inter-episode returns:
\begin{equation}
    g_{t}^{(n)} = \sum_{l=t}^{T^{(n)}-1} \gamma_{\text{step}}^{l-t}\, r_l^{(n)}, \qquad
    G_{t}^{(n)} =   g_{t}^{(n)} + \sum_{m=n+1}^{N-1} \gamma_{\text{traj}}^{m-n}\, g_0^{(m)}.
    \label{eq:dual_discount}
\end{equation}
Where $\gamma_{\text{step}}$ and $\gamma_{\text{traj}}$ are the intra- and inter-episode discount factors, respectively.

However, naively applying this in the pivotal setting lets erroneous steps ($t \geq k^{*(n)}$) receive positive returns from the successful retry, rewarding the very mistakes that caused failure. We introduce pivotal credit isolation to partition steps at $k^{*(n)}$:
\begin{equation}
    G_t^{(n)} = 
    \begin{cases}
        g_t^{(n)} + \gamma_{\text{traj}} \cdot G_0^{(n+1)}, & t < k^{(n)} \; \\[4pt]
        \displaystyle\sum_{l=t}^{T^{(n)}-1} \gamma_{\text{step}}^{l-t}\, r_l^{(n)}, & t \geq k^{(n)} \;
    \end{cases}
    \label{eq:pivotal_isolation}
\end{equation}

We compute $G_t^{(n)}$ backward from $\tau^{(N-1)}$ to $\tau^{(0)}$, so later retries serve as prior credit sources. Pivotal isolation prevents this future credit from leaking into the erroneous suffix $t\geq k^{*(n)}$.

\paragraph{Reflection Reward through Implicit Returns.} To co-optimize reflection generation and action retries, we define the reflection reward $r_{\text{reflect}}^{(n)}$ as the discounted return of the subsequent retry episode:
\begin{equation}
r_{\text{reflect}}^{(n)} = \mathbb{I}(\text{success}) \cdot \sum_{l=k^*}^{T^{\prime}-1} \gamma_{\text{step}}^{l-k^*} r_l^{(n+1)}
\end{equation}
This reward incentivizes precise failure localization by coupling reflection quality with retry success and penalizing redundancy via $\gamma_{\text{step}}$. Moreover, the reward is contingent upon correct formatting to ensure structural validity.

\paragraph{Optimization.} Given the per-action cross-episode return $G_t^{(n)}$ defined above, we define the gradient objective as follows:
\begin{equation}
\begin{aligned}
\nabla_{\theta} \mathcal{L}_{\text{RL}}(\theta)
&= \mathbb{E}_{\bm{\tau} \sim \pi_{\theta}}
\bigg[
\sum_{n,t}
\nabla_{\theta} \log \pi_{\theta}
\bigl(a_t^{(n)} \mid s_t^{(n)}\bigr)
A_t^{(n)} 
+
\sum_{n}
\nabla_{\theta} \log \pi_{\theta}
\bigl(\cdot \mid x,\bm{c}^{n}\bigr)
A_{\text{reflect}}^{(n)}
\bigg].
\end{aligned}
\label{eq:final}
\end{equation}

Where $A_t^{(n)}$ is the advantage estimate from $G_t^{(n)}$, and $A^{(n)}_{\text{reflect}}$ is derived via group-wise normalization of the reflection rewards $r^{(n)}_{\text{reflect}}$. The framework is compatible with widely used optimizers such as GRPO and GiGPO~\citep{gigpo}. See Appendix D for the detailed advantage computation.

Notably, pivotal retry also improves the model's first-attempt success ability (See in  Appendix~\ref{app:preference}).

\subsection{Concentrated Information Gain Analysis}
\label{sec:information_gain}

We analyze \ourapproach via an information-gain lens. Unlike global reflections facing dilution , \ourapproach concentrates guidance at the error boundary maximizing efficiency.

\paragraph{Per-Turn Information Gain.}
Given a trajectory $\tau$ and experience guidance $\textbf{remark}^{(n)}$, we define the \emph{per-turn information gain}:
\begin{equation}
    \mathrm{IG}(e, t) = \log \pi_\theta(a_t^* \mid s_t, e) - \log \pi_\theta(a_t^* \mid s_t).
    \label{eq:information_gain}
\end{equation}
High $\mathrm{IG}(e, t)$ indicates actionable guidance at turn $t$; near-zero values indicate noise.

We argue that existing experience utilization paradigms suffer from low \emph{effective signal density}~\citep{ragenv2}: the fraction of turns where the guidance produces non-negligible IG.

\begin{proposition}[IG degradation under state-agnostic guidance]
\label{prop:ig_degradation}
Let $\rho(e) = \frac{1}{T}|\{t : \mathrm{IG}(e, t) > \delta\}|$ denote the effective signal density of guidance $e$. For state-agnostic guidance---whether a fixed reflection $\Delta$ reused across all turns, or a retrieved experience $e^*$ from a finite memory bank---the per-turn IG satisfies $\mathrm{IG}(e, t) = O(1/T)$, yielding $\rho(e) \to 0$ as $T$ grows. Proof and detailed case analysis are provided in Appendix~\ref{app:ig_proofs}.
\end{proposition}

% \noindent Proof and detailed case analysis are provided in Appendix~\ref{app:ig_proofs}.

\paragraph{Concentrated Information Gain via Pivotal Retry.}
By identifying the error boundary $k^*$ and applying state-specific guidance at the pivotal turn, our mechanism concentrates information gain:
\begin{equation}
    \mathrm{IG}(e, t) \approx
    \begin{cases}
        \mathrm{IG}_{\mathrm{high}}, & t \ge  k^{*}, \\
        0, & t <  k^{*},
    \end{cases}
    \label{eq:ours_ig}
\end{equation}
where $\mathrm{IG}_{\mathrm{high}} \gg \frac{1}{T}\sum_t \mathrm{IG}(\Delta,t)$. 
Although the effective signal density is $\rho(e)=1/T$, the total IG is concentrated rather than diluted: the correct prefix requires no guidance, while the error suffix is regenerated from $k^*$. 
As training proceeds, $k^*$ moves later, yielding more precise experience signals.

\section{Experiments}
\label{sec:experiments}

% We evaluate \ourapproach on four challenging benchmarks for LLM agents. Our experiments address the following questions: 1) How does \ourapproach compare to stateof-the-art methods? 2) What is the contribution of each component? 3)How did the reflective behavior evolve during training?? 4) Does experience internalization accelerate model convergence?

\subsection{Experimental setup}
\label{sec:exp_setup}

\paragraph{Environment.}  We evaluate \ourapproach on four agent exploration  environments ( Sokoban~\citep{racaniere2017imagination}, Minesweeper~\citep{li2024assessing}, WebShop~\citep{yao2022webshop}, and ALFWorld~\citep{shridhar2020alfworld}) and multi-step search tasks, where we use a constructed Wiki knowledge base as the retrieval source. And we evaluate the multi-turn search performance on single-hop QA datasets (NQ~\citep{nqsearch}, TriviaQA~\citep{TriviaQA}, and PopQA~\citep{popqa}) and multi-hop QA datasets (HotpotQA~\citep{yang2018hotpotqa}, 2Wiki~\citep{xanh2020_2wikimultihop}, MuSiQue~\citep{trivedi2021musique}, and Bamboogle~\citep{Bamboogle}).  Details are provided in Appendix~\ref{app:details_of_exp}.

\paragraph{Baselines.}
We compare \ourapproach with six categories of competitive baselines. 
1) \textit{\textbf{Closed-source LLMs}}, including GPT-4o and Gemini-2.5-Pro. 
2) \textit{\textbf{Prompt-based Agentic Methods}}, including ReAct~\citep{yao2023react} and Reflexion~\citep{shinn2023reflexion}. 
3) \textit{\textbf{RL-based Methods}}, including group-based online RL algorithms such as GRPO and GiGPO, as well as the clipping-based RL algorithm GSPO~\citep{gspo}. 
4) \textit{\textbf{Memory-Augmented RL-based Methods}}, including Mem0+GRPO~\citep{mem0}, SimpleMem+GRPO~\citep{simplemem}, and SkillRL~\citep{xia2026skillrlevolvingagentsrecursive}. 
5) \textit{\textbf{Reflect-Retry RL-based Methods}}, including MetaRL~\citep{lamer}. 
6) \textit{\textbf{Classic Agentic Search Methods}}, including Search-R1~\citep{searchr1}, ZeroSearch~\citep{zerosearch}, and StepSearch~\citep{stepsearch}.

\paragraph{Training details.}  We use Qwen3-4B as the base model and adopt GiGPO as the default optimization algorithm across all experiments. For \ourapproach training, we sample $N=3$ episodes and set the group size to 8 for agent tasks and group size to 4 for search tasks. All experiments are conducted on a cluster with 8 NVIDIA H100-80GB GPUs. Detailed training settings are provided in Appendix~\ref{app:details_of_exp}.

\begin{table*}[t]
\caption{Main results across four agent benchmarks with Qwen3-4B. \textbf{Suc@1/2/3}: success rate in 1, 2 or 3 rollouts (\%);  \textbf{Avg.} denotes the mean of all success metrics across environments. The best results and second best results are highlighted in \colorbox{red!25}{red} and \colorbox{blue!15}{blue}, respectively.}
\label{tab:main_results}
\centering
\setlength{\tabcolsep}{2.5pt}
\renewcommand{\arraystretch}{1.05}
\resizebox{\textwidth}{!}{%
\footnotesize
\begin{tabular}{@{}l ccc ccc ccc ccc >{\columncolor{orange!8}}c @{}}
\toprule
\multirow{2}{*}{\textbf{Methods}} 
& \multicolumn{3}{c}{\textbf{ALFWorld}} 
& \multicolumn{3}{c}{\textbf{WebShop}} 
& \multicolumn{3}{c}{\textbf{Minesweeper}} 
& \multicolumn{3}{c}{\textbf{SciWorld}} 
& \multirow{2}{*}{\textbf{Avg.}} \\
\cmidrule(lr){2-4} 
\cmidrule(lr){5-7} 
\cmidrule(lr){8-10} 
\cmidrule(lr){11-13}
& Suc@1 & Suc@2 & Suc@3
& Suc@1 & Suc@2 & Suc@3 
& Suc@1 & Suc@2 & Suc@3 
& Suc@1 & Suc@2 & Suc@3
& \\
\midrule

\rowcolor{gray!15} 
\multicolumn{14}{@{}l}{\textit{\textbf{Closed-source LLMs}}} \\
\quad GPT-4o             
& 42.5 & 48.3 & 50.8
& 33.6 & 37.8 & 39.5
& 22.5 & 26.4 & 26.7
& 32.8 & 42.1 & 47.6 
& 37.6 \\
\quad Gemini-2.5-Pro    
& 54.3 & 61.2 & 64.8
& 37.6 & 39.7 & 40.4
& 24.2 & 29.6 & 31.2 
& 42.9 & 55.4 & 60.9  
& 45.2 \\

\midrule
\rowcolor{gray!15} 
\multicolumn{14}{@{}l}{\textit{\textbf{Prompt-based Agentic Methods}}} \\
\quad ReAct         
& 17.1 & 21.8 & 27.2
& 2.30 & 5.40 & 7.5 
& 3.12 & 3.96 & 5.83 
& 9.4 & 12.5 & 12.5 
& 10.7 \\
\quad Reflexion    
& 17.9 & 28.1 & 29.7
& 1.55 & 3.18 & 5.1 
& 3.12 & 5.46 & 7.49 
& 9.3 & 14.7 & 17.9
& 11.9 \\

\midrule
\rowcolor{gray!15} 
\multicolumn{14}{@{}l}{\textit{\textbf{RL-based Methods}}} \\
\quad GRPO         
& 76.4 & 80.3 & 82.6 
& 61.7 & 68.8 & 72.7 
& 39.1 & 42.1 & 45.3 
& 75.8 & 82.8 & 86.7 
& 67.9 \\
\quad DAPO         
& 82.0 & 89.8 & 91.2 
& 64.1 & 69.5 & 75.0 
& 46.0 & 50.0 & 51.6 
& 83.5 & 94.5 & \colorbox{blue!15}{98.4} 
& 74.6 \\
\quad GiGPO        
& 85.9 & 87.5 & 89.8 
& \colorbox{blue!15}{67.9} & \colorbox{blue!15}{80.5} & \colorbox{blue!15}{82.6} 
& 42.2 & 46.1 & 48.4 
& 78.9 & 86.7 & 89.8
& 73.9 \\
\quad GSPO         
& 85.9 & \colorbox{blue!15}{91.4} & 91.4
& 61.7 & 67.2 & 68.6
& 46.1 & \colorbox{blue!15}{57.0} & 57.8 
& 89.8 & 94.5 & 96.0 
& 75.6 \\

\midrule
\rowcolor{gray!15} 
\multicolumn{14}{@{}l}{\textit{\textbf{Memory-Augmented RL-based Methods}}} \\
\quad Mem0+GRPO        
& 82.8 & 84.7 & 85.2 
& 63.1 & 69.5 & 74.8
& 44.5 & 46.8 & 49.2
& 79.6 & 91.4 & 93.2 
& 72.1 \\
\quad SimpleMem+GRPO        
& \colorbox{blue!15}{87.5} & 89.0 & 90.6 
& 66.4 & 74.2 & 78.1 
& 43.0 & 50.0 & 51.5 
& 81.7 & 90.2 & 92.6
& 74.6 \\
\quad SKillRL        
& 83.6 & \colorbox{blue!15}{91.4} & \colorbox{blue!15}{92.5} 
& \colorbox{red!25}{\textbf{68.2}} & 72.6 & 74.2
& \colorbox{red!25}{\textbf{50.7}} & 56.3 & 57.9
& 82.1 & 89.0 & 89.8 
& 75.7 \\

\midrule
\rowcolor{gray!15} 
\multicolumn{14}{@{}l}{\textit{\textbf{Reflect-Retry RL-based Methods}}} \\
\quad MetaRL 
& 82.0 & 85.9 & 86.7 
& 63.2 & 71.3 & 75.2 
& 39.8 & 54.9 & \colorbox{blue!15}{61.5} 
& \colorbox{blue!15}{94.5} & \colorbox{blue!15}{97.6} & \colorbox{blue!15}{98.4}
& \colorbox{blue!15}{75.9} \\

\rowcolor{blue!5}
\quad \textbf{\ourapproach (Ours) }
& \colorbox{red!25}{\textbf{89.8}} & \colorbox{red!25}{\textbf{93.4}} & \colorbox{red!25}{\textbf{94.5}} 
& 66.7 & \colorbox{red!25}{\textbf{81.5}} & \colorbox{red!25}{\textbf{83.8}} 
& \colorbox{blue!15}{48.4} & \colorbox{red!25}{\textbf{71.8}} & \colorbox{red!25}{\textbf{78.1}}
& \colorbox{red!25}{\textbf{98.4}} & \colorbox{red!25}{\textbf{100}} & \colorbox{red!25}{\textbf{100}} 
& \colorbox{red!25}{\textbf{83.9}} \\

\bottomrule
\end{tabular}%
}
\end{table*}
\begin{table}[t]
\centering
\caption{Pass@3 performance on search-augmented QA tasks with Qwen3-4B. Models are trained on NQ and HotpotQA. $\dagger$ and $\star$ indicate in-domain and out-of-domain datasets, respectively. The best results and second best results are highlighted in \colorbox{red!25}{red} and \colorbox{blue!15}{blue}, respectively.}
\label{tab:main_qa}
\resizebox{\textwidth}{!}{
\begin{tabular}{lcc@{\,\,\,}c|ccc@{\,\,\,}c|c}
\toprule
\multirow{2}{*}{Method} & \multicolumn{3}{c|}{\textbf{Single-Hop QA}} & \multicolumn{4}{c|}{\textbf{Multi-Hop QA}} \\
& NQ$^{\dagger}$ & TriviaQA$^{\star}$ & PopQA$^{\star}$ & HotpotQA$^{\dagger}$ & 2Wiki$^{\star}$ & MuSiQue$^{\star}$ & Bamboogle$^{\star}$ & Avg.\\
\midrule
\rowcolor{gray!15} 
\multicolumn{4}{l|}{\textit{\textbf{Classic Agentic Search Methods}}} 
& \multicolumn{4}{l|}{} 
& \multicolumn{1}{l}{} \\
Search-R1  
& 48.3 & 63.6 & 48.9 
& 47.3 & 52.4 & 22.9 & 46.4 
& 47.1 \\
ZeroSearch  
& 51.2 & 63.1 & 48.3 
& 48.1 & 47.3 & 23.1 & 46.4 
& 46.8 \\
StepSearch 
& 49.3 & 65.8 & 47.9 
& 48.8 & 50.0 & 22.4 & 51.2 
& 47.9 \\
\midrule
\rowcolor{gray!15} 
\multicolumn{4}{l|}{\textit{\textbf{RL-based Methods}}} 
& \multicolumn{4}{l|}{} 
& \multicolumn{1}{l}{} \\
GRPO  
& 49.7 & 63.7 & 50.0
& 47.3 & 46.4 & 20.7 & 47.2 
& 46.4 \\
GiGPO  
& 49.7 & 67.2 & 49.1 
& 48.2 & 47.9 & 20.0 & 48.0 
& 47.2 \\
GSPO   
& 48.8 & 64.8 & \colorbox{blue!15}{50.6} 
& 45.9  & 43.0 & 18.5 & 40.8 
& 44.6 \\
\midrule
\rowcolor{gray!15} 
\multicolumn{4}{l|}{\textit{\textbf{Reflect-Retry RL-based Methods}}}  
& \multicolumn{4}{l|}{} 
& \multicolumn{1}{l}{} \\
MetaRL   
& \colorbox{blue!15}{51.7} & \colorbox{blue!15}{67.8} & 49.8 
& \colorbox{red!25}{\textbf{54.6}} & \colorbox{blue!15}{55.2} & \colorbox{blue!15}{24.4} & \colorbox{blue!15}{53.6} 
& \colorbox{blue!15}{51.0} \\
\rowcolor{blue!5} \textbf{\ourapproach (Ours)}  
& \colorbox{red!25}{\textbf{59.8}} & \colorbox{red!25}{\textbf{71.6}} & \colorbox{red!25}{\textbf{59.4}} 
& \colorbox{blue!15}{53.0} & \colorbox{red!25}{\textbf{64.0}} & \colorbox{red!25}{\textbf{31.1}} & \colorbox{red!25}{\textbf{59.2}} 
& \colorbox{red!25}{\textbf{56.9}} \\
\bottomrule
\end{tabular}
}
\end{table}

\subsection{Main Result}
\label{sec:result}

\begin{figure*}[t]
\centering
\includegraphics[width=\textwidth]{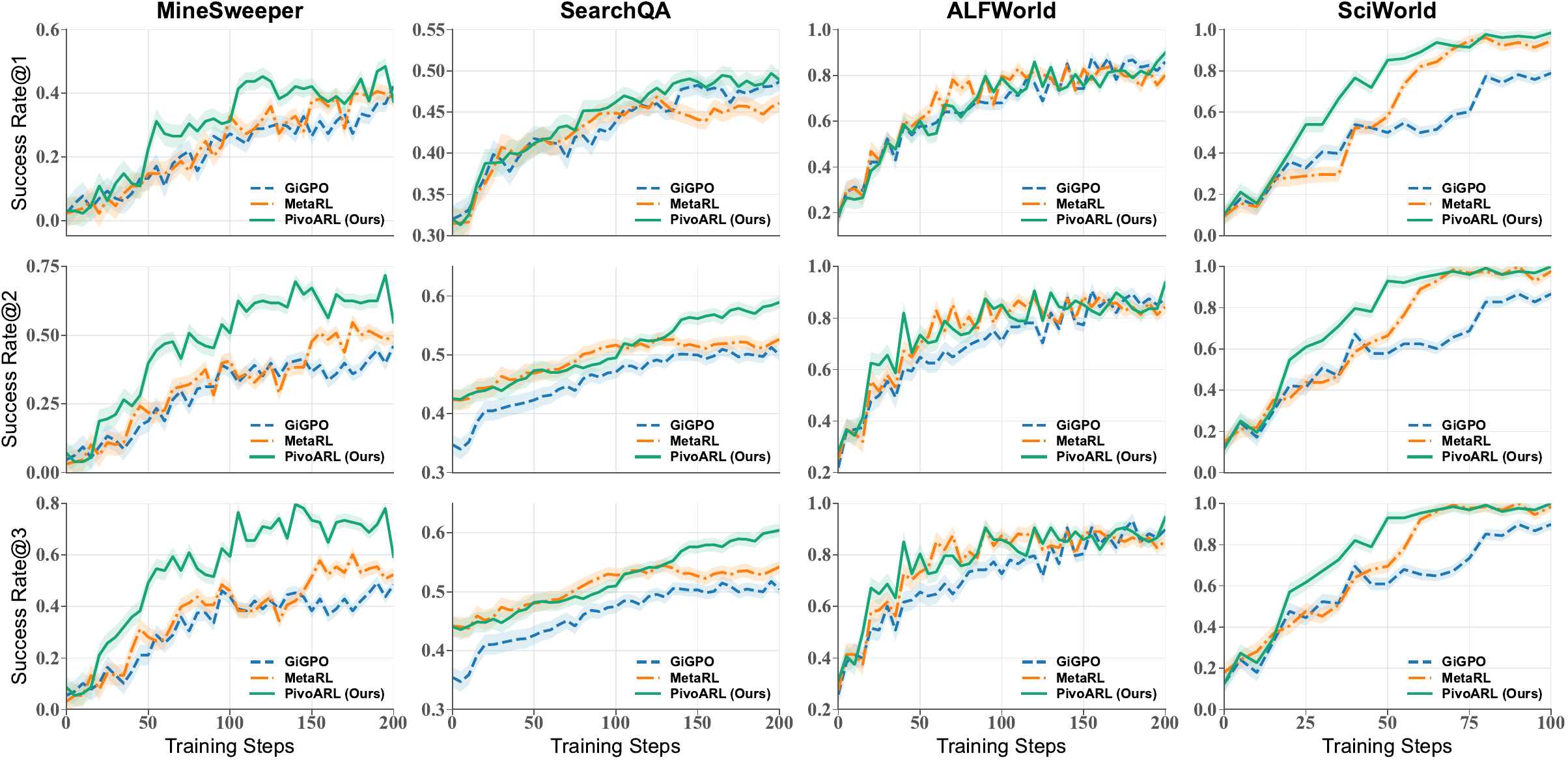}
\caption{Training curves across four agent benchmarks on Qwen3-4B. The top, middle, and bottom rows show Success@1, Success@2, and Success@3, respectively, plotted against training steps. We compare GiGPO (gray, dotted), MetaRL (blue, dashed), \ourapproach (blue, dashed).}
\label{fig:training_curves}
\end{figure*}

We evaluate all methods on four agent benchmarks and seven search-based QA benchmarks. Table~\ref{tab:main_results} summarizes the final performance on agent tasks, while Table~\ref{tab:main_qa} reports comparisons with search-related methods. Figure~\ref{fig:training_curves} shows the training curves across all four tasks.

\paragraph{Comparison on Agent Tasks.} Across all four agent environments, \ourapproach consistently outperforms other RL training methods and Reflect-Retry baselines on all Succ@2/3 metrics. Overall, our method achieves an average 10.5\% improvement over the state-of-the-art MetaRL. In the Minesweeper environment, \ourapproach surpasses GIGPO and MetaRL by 27.7\% and 16.8\%, respectively. Moreover, benefiting from contrastive advantages at pivotal turns, \ourapproach also obtains an average 6.0\% gain on most Succ@1 metrics, demonstrating its superior effectiveness.

\textbf{Comparison on Search Tasks.} Across seven single-hop and multi-hop retrieval tasks, \ourapproach achieves the best Pass@3 performance on 6/7 benchmarks, and obtains the second-best result on HotpotQA. Compared with the GRPO baseline, \ourapproach obtains a 22.6\% relative improvement. Compared with classic agentic search methods, \ourapproach improves performance by 20.4\% on average. Compared with the reflect-retry baseline MetaRL, \ourapproach further achieves an 11.6\% relative improvement. Detailed Pass@1 and Pass@2 results are provided in Appendix E.

\paragraph{Comparison with Training Dynamics.} We present the Pass@1/2/3 training curves of \ourapproach, GiGPO, and MetaRL. Overall, \ourapproach shows more stable training advantages across different tasks. On Minesweeper, \ourapproach significantly accelerates learning convergence and consistently improves all three metrics. On search tasks, \ourapproach also brings moderate gains in Pass@1, with more pronounced improvements on Pass@2/3, indicating stronger advantages in multi-attempt settings.

\subsection{Ablation studies}
\label{sec:ablations}
\begin{wraptable}{r}{0.52\textwidth}
\vspace{-12pt}
\centering
\caption{Ablation Study of \ourapproach component. }
\label{tab:ablation}
\setlength{\tabcolsep}{3pt}
\footnotesize
\begin{tabular}{@{}lccc@{}}
\toprule
\textbf{Method} & \textbf{ALFWorld} & \textbf{SciWorld} & \textbf{Minesweeper} \\
\midrule
\rowcolor[RGB]{236,244,252}
\textbf{GiGPO} & 89.8 & 89.8 & 48.4 \\
\midrule
\textit{\ourapproach} & & & \\
\quad w/o \textit{Reflection} & 87.5 & 86.3 & 50.5 \\
\quad w/o \textit{Pivotal}  & 86.7 &  98.4 & 61.5  \\
\quad w/o \textit{Isolation}  & 91.8 & 95.9 & 54.6  \\
\midrule
\rowcolor[RGB]{236,244,252}
\textbf{\ourapproach} & \textbf{94.5} & \textbf{100} & \textbf{78.1} \\
\bottomrule
\end{tabular}
\vspace{-8pt}
\end{wraptable}

\paragraph{Ablation on Component.} To analyze the contribution of each component in \ourapproach, we conduct the following ablation experiments: a) \textit{\textbf{w/o Reflection}}: removing reflection experience injection; b) \textit{\textbf{w/o Pivotal}}: removing pivotal turn localization and instead performing a full retry from the initial state; and c) \textit{\textbf{w/o Isolation}}: removing the pivotal-isolated credit assignment mechanism. The results are shown in Table~\ref{tab:ablation}. Without reflection experience, the model performs similarly to vanilla GiGPO, indicating that reflection injection is an important source of improved exploration efficiency. Removing the pivotal mechanism reduces our method to a full-retry paradigm similar to MetaRL. In contrast, credit isolation plays a key role in pivotal retry by preventing gradient updates from mistakenly penalizing the correct prefix, bringing an average improvement of 10.1\% percentage points over the w/o Isolation, with a particularly large gain of 23.5\% percentage points on Minesweeper.

\paragraph{Ablation on Credit Assignment.} As shown in Figure~\ref{fig:ablation}, \ourapproach demonstrates superior exploration efficiency and prompt economy over baselines. In terms of reflection capability (Fig.~\ref{fig:ablation}a), \ourapproach achieves a peak success rate of  83.8\%, outperforming MetaRL 58.9\% and the baseline. When evaluating retry efficiency (Fig.~\ref{fig:ablation}b), our method maintains a competitive average retry ratio of  0.44, requiring significantly fewer redundant interactions than MetaRL (constant 1.0). Notably, removing the CAPI mechanism (w/o CAPI) causes a noticeable drop in success rate and an increase in retry frequency, confirming CAPI's necessity. Furthermore, analysis of token overhead (Fig.~\ref{fig:ablation}c) reveals that \ourapproach consumes only  716 additional guidance tokens on average—comparable to the ablated variant and substantially fewer than the baseline. 

%This confirms that \ourapproach achieves enhanced reasoning performance without imposing excessive context window burdens.

\subsection{Analysis}
\label{sec:analysis}

\paragraph{Analysis of Preference Group Density.} To understand why pivotal retry also improves Pass@1, we count the preference groups produced by different retry strategies. A preference group refers to trajectories that share the same state but receive different rewards. As shown in Figure~\ref{fig7:preference},  reflective retry  consistently produces the highest number of preference groups throughout the training process, achieving an overall 48.4\% increase over MetaRL, mainly concentrated in the middle stages of trajectories. This indicates that it provides denser optimization signals around critical decision regions, leading to stable Pass@1 improvement. In contrast, random retry performs worse than full retry (MetaRL), suggesting that the gain comes from structured reflection.

\begin{wrapfigure}[10]{r}{7cm}
\centering
\vspace{-5mm}
\includegraphics[width=0.5\textwidth]{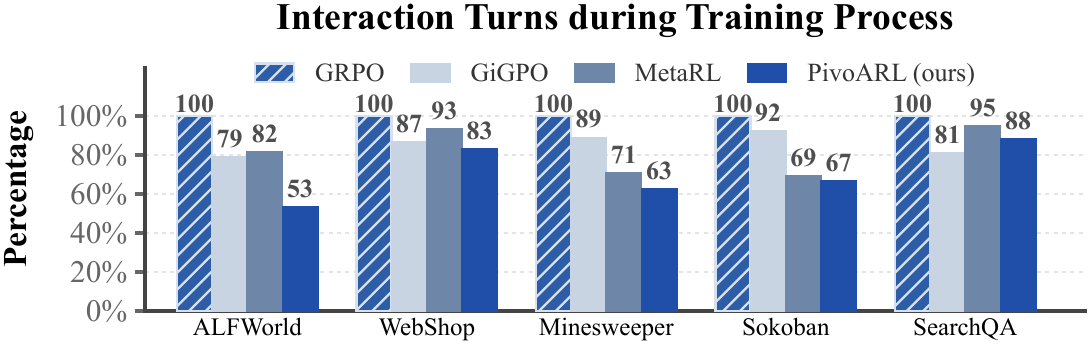}
\vspace{-5mm}
\caption{\label{fig9:interact} Comparison of interaction counts across training algorithms, with GRPO set to 100\%. }
\end{wrapfigure}
\paragraph{Analysis of Training Interaction Cost.} PivoARL reuses the correct prefix through pivotal retry, avoiding redundant interactions caused by full retries. To evaluate its interaction efficiency, we compare PivoARL with GRPO, GiGPO, and MetaRL, as shown in Figure~\ref{fig9:interact}. We normalize the number of training interactions by setting GRPO to 100\%. Results show that PivoARL reduces the required interactions across all tasks, lowering interaction cost by about 29.1\% on average compared with GRPO and by about 13.5\% compared with MetaRL. This demonstrates that PivoARL improves performance while offering higher training efficiency.

\begin{figure*}[t]
\centering
\includegraphics[width=\textwidth]{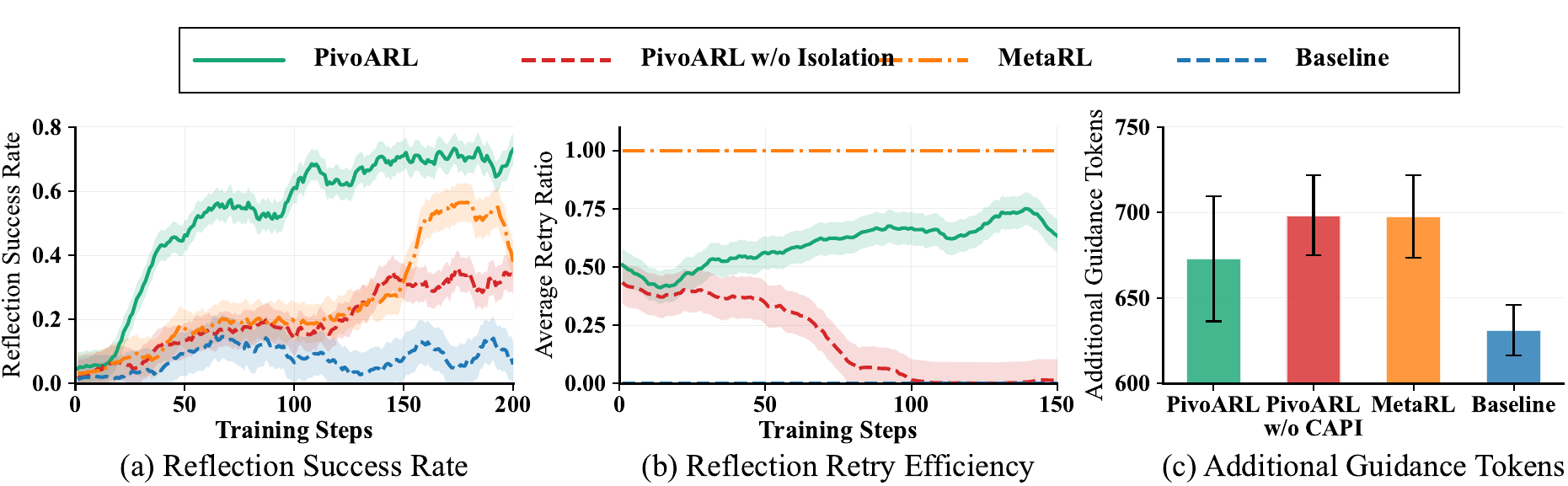}
\caption{Performance and efficiency comparison on MineSweeper. We evaluate \ourapproach, its ablated variant (w/o Credit Assignment with Pivotal Isolation (CAPI)), MetaRL, and the Baseline across three metrics: (a) Reflection Success Rate, (b) Average Retry Ratio, and (c) Prompt Length.}
\label{fig:ablation}
\end{figure*}

\paragraph{Exhibits Stronger Test-time Scaling.} To evaluate whether \ourapproach can better utilize additional test-time attempts, we measure Pass@k by varying the number of rollouts during evaluation. As shown in Figure~\ref{fig7:test}, \ourapproach consistently outperforms both standard RL and full-retry reflection baselines across all values of k. Averaged over Minesweeper and SearchQA, \ourapproach improves from 47.0\% Pass@1 to 74.7\% Pass@8, while MetaRL improves from 41.4\% to 62.4\%. These results indicate that \ourapproach exhibits stronger test-time scaling by learning to use additional attempts more effectively through reflective error localization.

\paragraph{Scaling Performance with Episode Nums.} We further study the training-time scaling behavior of \ourapproach by varying the number of training episodes (N). As shown in Figure~\ref{fig7:train}, \ourapproach generally achieves stronger Pass@k performance as (N) increases. In particular, Pass@1 improves from 39.8\% at (N=1) to 53.9\% at (N=5), while the best overall result reaches 88.4\% under the largest episode budget. These results suggest that increasing the number of training episodes provides richer retry and preference signals, enabling \ourapproach to scale effectively during training.

\section{Related Work}
\label{sec:related_work}

\paragraph{Agent Reinforcement Learning.}
RL plays a crucial role in helping LLM agents adapt to dynamic and open-ended environments.
Recent work such as RAGEN~\citep{wang2025ragen} integrates reasoning states with environment interactions into turn-level responses and trains with trajectory-level RL.
To address sparse rewards in multi-turn interactions, IGPO~\citep{igpo} introduces information-gain-based intrinsic rewards derived from the model's own confidence.
To better capture multi-turn agent behaviors that trajectory-level RL struggles with, Tree-GRPO~\citep{treegrpo} employs tree-search-based group relative policy optimization to enable learning from intermediate decisions. In addition, GSPO~\citep{gspo} CISPO~\citep{cispo} and SAPO~\citep{sapo} focus on the stability of policy optimization, mitigating high-variance updates through different importance-ratio clipping mechanisms. However, these methods treat each interaction as an isolated event, making it difficult to leverage past trajectory experience for training.

\paragraph{Self-Feedback Evolving Agents.}
Recently, a growing number of methods have begun to emphasize the role of interaction experience in the continual evolution of agents.
\emph{In-context learning} approaches such as Reflect-GRPO~\citep{reflectgrpo} and RLTF~\citep{rltf} treat natural language reflections as instructive experience~\citep{shinn2023reflexion,mage,mrsearch}.
To internalize such short-term experience into the model's inherent capabilities, the \emph{offline distillation} line of work---including ERL~\citep{erl}, R$^3$L~\citep{r3l}, and Critique-GRPO~\citep{critiquegrpo} uses corrected trajectories or critiques to guide offline optimization. Furthermore, self-distillation, represented by SDAR~\citep{sdar} and SDPO~\citep{sdpo}, turns the model's in-context learning ability into a ``self-teacher,'' internalizing feedback signals into parameters. MetaRL~\citep{lamer} and MR-Search~\citep{mrsearch} view a trajectory as a continuous learning process across multiple retry episodes, enabling cross-episode experience utilization.

\begin{figure}[t] % Use figure* to span both columns in a two-column layout; use figure for a single-column layout.
    \centering
    % First subfigure
    \begin{subfigure}{0.32\linewidth}
        \centering
        \includegraphics[width=\linewidth]{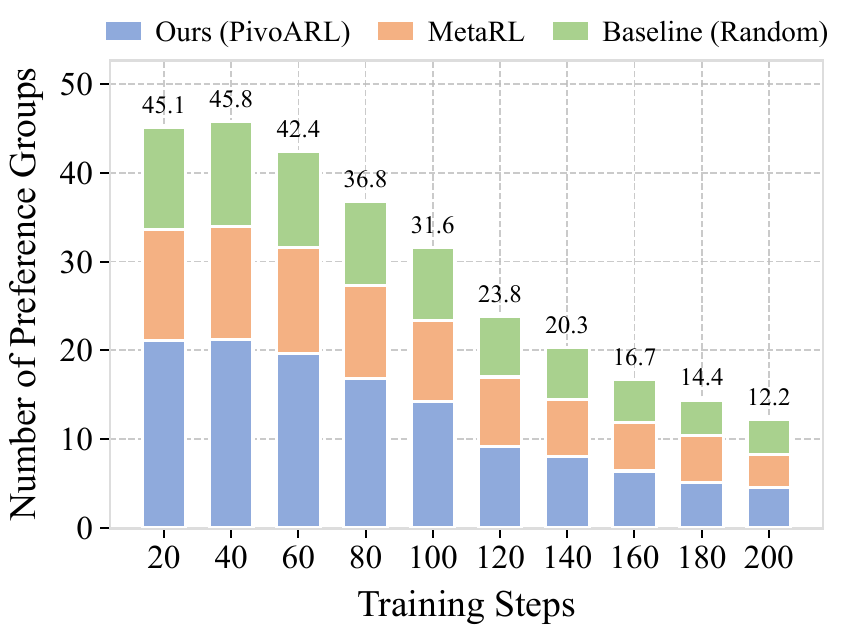}
        \caption{Change of Preference groups}
        \label{fig7:preference}
    \end{subfigure}
    \hfill % Inserts elastic horizontal spacing between subfigures
    % Second subfigure
    \begin{subfigure}{0.32\linewidth}
        \centering
        \includegraphics[width=\linewidth]{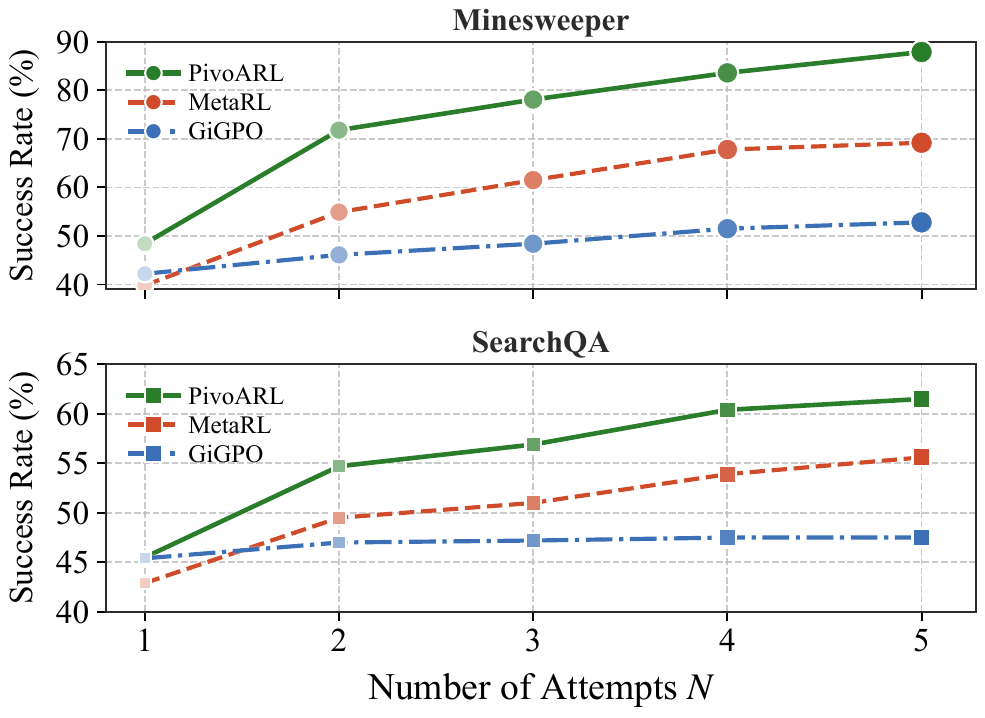}
        \caption{Test-time Scaling}
        \label{fig7:test}
    \end{subfigure}
    \hfill
    % Third subfigure
    \begin{subfigure}{0.32\linewidth}
        \centering
        % 同时指定 width 和 height，不使用 keepaspectratio
        \includegraphics[width=\linewidth]{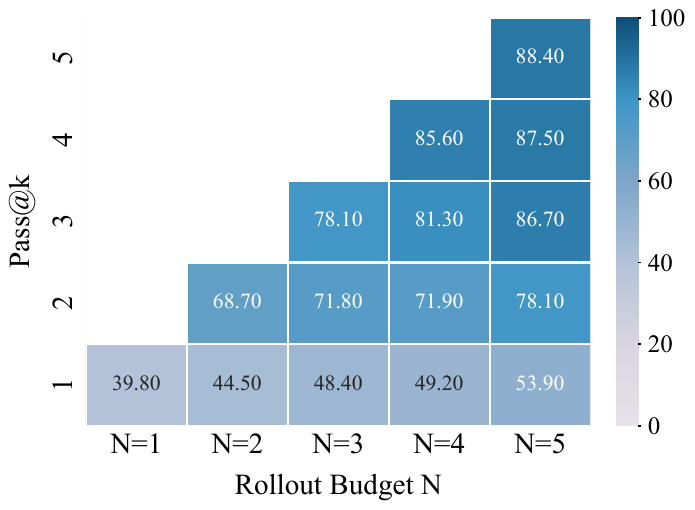}
        \caption{Train-time Scaling}
        \label{fig7:train}
    \end{subfigure}
    \caption{Analysis of the \ourapproach mechanism. We examine its impact through (a) dynamic preference-group counts, (b) test-time scalability, and (c) scaling with training episodes.}
    \label{fig:analyse_success}
\end{figure}

\section{Conclusion}
\label{sec:conclusion}

We proposed \ourapproach, a self-feedback reinforcement learning framework for exploiting failed trajectories in LLM agents. Instead of restarting from scratch, \ourapproach locates the pivotal erroneous turn and retries from that state, thereby reusing correct prefixes and reducing redundant interactions. This design concentrates experience signals around critical error boundaries and prevents erroneous suffixes from receiving misleading credit. Extensive experiments show that \ourapproach consistently improves agent performance while reducing interaction cost.

\paragraph{Limitation and Future Work.}
\ourapproach depends on accurate pivotal-turn localization, and incorrect reflection may reduce retry effectiveness. It also introduces extra training cost compared with single-rollout RL. Future work may explore more reliable pivotal-state verification, adaptive retry allocation, and extensions to more complex real-world tool-use agents.

\bibliography{iclr2026_conference}

\begin{thebibliography}{43}
\providecommand{\natexlab}[1]{#1}
\providecommand{\url}[1]{\texttt{#1}}
\expandafter\ifx\csname urlstyle\endcsname\relax
  \providecommand{\doi}[1]{doi: #1}\else
  \providecommand{\doi}{doi: \begingroup \urlstyle{rm}\Url}\fi

\bibitem[Bensal et~al.(2025)Bensal, Jamil, Bryant, Russak, Kamble, et~al.]{reflectgrpo}
Shelly Bensal, Umar Jamil, Christopher Bryant, Melisa Russak, Kiran Kamble, et~al.
\newblock Reflect, retry, reward: Self-improving {LLM}s via reinforcement learning.
\newblock \emph{arXiv preprint arXiv:2505.24726}, 2025.

\bibitem[Chhikara et~al.(2025)Chhikara, Khant, Aryan, Singh, and Yadav]{mem0}
Prateek Chhikara, Dev Khant, Saket Aryan, Taranjeet Singh, and Deshraj Yadav.
\newblock Mem0: Building production-ready {AI} agents with scalable long-term memory.
\newblock In \emph{Proceedings of 28th European Conference on Artificial Intelligence (EACL)}, 2025.

\bibitem[Feng et~al.(2025)Feng, Xue, Liu, and An]{gigpo}
Lang Feng, Zhenghai Xue, Tingcong Liu, and Bo~An.
\newblock Group-in-group policy optimization for {LLM} agent training.
\newblock In \emph{Proceedings of the Advances in Neural Information Processing Systems (NeurIPS)}, 2025.

\bibitem[Gao et~al.(2025)Gao, Zheng, Chen, Dang, Liu, Yu, Yang, Bai, Zhou, and Lin]{sapo}
Chang Gao, Chujie Zheng, Xiong-Hui Chen, Kai Dang, Shixuan Liu, Bowen Yu, An~Yang, Shuai Bai, Jingren Zhou, and Junyang Lin.
\newblock Soft adaptive policy optimization.
\newblock \emph{arXiv preprint arXiv:2511.20347}, 2025.

\bibitem[Guo et~al.(2025)Guo, Li, Wang, Li, He, Yu, and Zhang]{guo2025mtsa}
Weiyang Guo, Jing Li, Wenya Wang, Yu~Li, Daojing He, Jun Yu, and Min Zhang.
\newblock Mtsa: Multi-turn safety alignment for llms through multi-round red-teaming.
\newblock In \emph{Proceedings of the 63rd Annual Meeting of the Association for Computational Linguistics (Volume 1: Long Papers)}, pp.\  26424--26442, 2025.

\bibitem[Guo et~al.(2026)Guo, Shi, Zhao, Ma, Zhu, He, Zhang, and Li]{guo2026e3tirenhancedexperienceexploitation}
Weiyang Guo, Zesheng Shi, Liye Zhao, Jiayuan Ma, Zeen Zhu, Junxian He, Min Zhang, and Jing Li.
\newblock E3-tir: Enhanced experience exploitation for tool-integrated reasoning.
\newblock 2026.

\bibitem[Ho et~al.(2020)Ho, Nguyen, Sugawara, and Aizawa]{xanh2020_2wikimultihop}
Xanh Ho, Anh-Khoa~Duong Nguyen, Saku Sugawara, and Akiko Aizawa.
\newblock Constructing a multi-hop {QA} dataset for comprehensive evaluation of reasoning steps.
\newblock In \emph{Proceedings of the International Conference on Computational Linguistics (COLING)}, pp.\  6609--6625, 2020.

\bibitem[H{\"u}botter et~al.(2026)H{\"u}botter, L{\"u}beck, Behric, Baumann, Bagatella, Marta, Hakimi, Shenfeld, Kleine~Buening, Guestrin, and Krause]{sdpo}
Jonas H{\"u}botter, Frederike L{\"u}beck, Lejs Behric, Anton Baumann, Marco Bagatella, Daniel Marta, Ido Hakimi, Idan Shenfeld, Thomas Kleine~Buening, Carlos Guestrin, and Andreas Krause.
\newblock Reinforcement learning via self-distillation.
\newblock \emph{arXiv preprint arXiv:2601.20802}, 2026.

\bibitem[Ji et~al.(2025)Ji, Ma, Wang, Chen, Chu, and Wu]{treegrpo}
Yuxiang Ji, Ziyu Ma, Yong Wang, Guanhua Chen, Xiangxiang Chu, and Liaoni Wu.
\newblock Tree search for llm agent reinforcement learning.
\newblock In \emph{Proceedings of the International Conference on Learning Representations (ICLR)}, 2025.

\bibitem[Jiang et~al.(2026)Jiang, Jiang, Teney, Moor, and Brbi{\'c}]{lamer}
Yulun Jiang, Liangze Jiang, Damien Teney, Michael Moor, and Maria Brbi{\'c}.
\newblock Meta-{RL} induces exploration in language agents.
\newblock In \emph{Proceedings of the International Conference on Learning Representations (ICLR)}, 2026.

\bibitem[Jin et~al.(2025)Jin, Zeng, Yue, Wang, Zamani, and Han]{searchr1}
Bowen Jin, Hansi Zeng, Zhenrui Yue, Dong Wang, Hamed Zamani, and Jiawei Han.
\newblock Search-r1: Training llms to reason and leverage search engines with reinforcement learning.
\newblock \emph{arXiv preprint arXiv:2503.09516}, 2025.

\bibitem[Joshi et~al.(2017)Joshi, Choi, Weld, and Zettlemoyer]{TriviaQA}
Mandar Joshi, Eunsol Choi, Daniel~S. Weld, and Luke Zettlemoyer.
\newblock Triviaqa: {A} large scale distantly supervised challenge dataset for reading comprehension.
\newblock In \emph{Proceedings of the 55th Annual Meeting of the Association for Computational Linguistics (ACL)}, 2017.

\bibitem[Kwiatkowski et~al.(2019)Kwiatkowski, Palomaki, Redfield, Collins, Parikh, Alberti, Epstein, Polosukhin, Kelcey, Devlin, Lee, Toutanova, Jones, Chang, Dai, Uszkoreit, Le, and Petrov]{nqsearch}
Tom Kwiatkowski, Jennimaria Palomaki, Olivia Redfield, Michael Collins, Ankur Parikh, Chris Alberti, Danielle Epstein, Illia Polosukhin, Matthew Kelcey, Jacob Devlin, Kenton Lee, Kristina~N. Toutanova, Llion Jones, Ming-Wei Chang, Andrew Dai, Jakob Uszkoreit, Quoc Le, and Slav Petrov.
\newblock Natural questions: a benchmark for question answering research.
\newblock \emph{Transactions of the Association for Computational Linguistics}, 2019.

\bibitem[Li et~al.(2026)Li, Zhang, Wang, Liu, Liang, Qi, Huang, Guo, Bao, Zhou, Mo, Xu, and Wu]{li2026mol}
Guocong Li, Jinjian Zhang, Ping Wang, Dongnan Liu, Tian Liang, Qiuyi Qi, Hao Huang, Siyan Guo, Mutian Bao, Wei Zhou, Linjian Mo, Hongxia Xu, and Jian Wu.
\newblock Mol: Adaptive mixture-of-length reasoning for efficient question answering with context.
\newblock In \emph{The Fourteenth International Conference on Learning Representations}, 2026.
\newblock URL \url{https://openreview.net/forum?id=oWWAeLEdE3}.

\bibitem[Li et~al.(2024)Li, Wang, and Zhang]{li2024assessing}
Yinghao Li, Haorui Wang, and Chao Zhang.
\newblock Assessing logical puzzle solving in large language models: Insights from a minesweeper case study.
\newblock In \emph{Proceedings of the 2024 Conference of the North American Chapter of the Association for Computational Linguistics (ACL)}, 2024.

\bibitem[Liu et~al.(2026)Liu, Su, Xia, Han, Zheng, Xie, Ding, and Yao]{simplemem}
Jiaqi Liu, Yaofeng Su, Peng Xia, Siwei Han, Zeyu Zheng, Cihang Xie, Mingyu Ding, and Huaxiu Yao.
\newblock Simplemem: Efficient lifelong memory for {LLM} agents.
\newblock \emph{arXiv preprint arXiv:2601.02553}, 2026.

\bibitem[Lu et~al.(2026)Lu, Yao, Han, Wang, Wu, Gu, Cai, Lu, Xiao, Zhuang, and Shen]{sdar}
Zhengxi Lu, Zhiyuan Yao, Zhuowen Han, Zi-Han Wang, Jinyang Wu, Qi~Gu, Xunliang Cai, Weiming Lu, Jun Xiao, Yueting Zhuang, and Yongliang Shen.
\newblock Self-distilled agentic reinforcement learning.
\newblock \emph{arXiv preprint arXiv:2605.15155}, 2026.

\bibitem[Mallen et~al.(2023)Mallen, Asai, Zhong, Das, Khashabi, and Hajishirzi]{popqa}
Alex Mallen, Akari Asai, Victor Zhong, Rajarshi Das, Daniel Khashabi, and Hannaneh Hajishirzi.
\newblock When not to trust language models: Investigating effectiveness of parametric and non-parametric memories.
\newblock In \emph{Proceedings of the 61st Annual Meeting of the Association for Computational Linguistics (ACL)}, 2023.

\bibitem[MiniMax et~al.(2025)MiniMax, :, Chen, Li, Gong, et~al.]{cispo}
MiniMax, :, Aili Chen, Aonian Li, Bangwei Gong, et~al.
\newblock Minimax-m1: Scaling test-time compute efficiently with lightning attention.
\newblock \emph{arXiv preprint arXiv:2506.13585}, 2025.

\bibitem[Press et~al.(2023)Press, Zhang, Min, Schmidt, Smith, and Lewis]{Bamboogle}
Ofir Press, Muru Zhang, Sewon Min, Ludwig Schmidt, Noah~A. Smith, and Mike Lewis.
\newblock Measuring and narrowing the compositionality gap in language models.
\newblock In \emph{Findings of the Association for Computational Linguistics (EMNLP)}, pp.\  5687--5711, 2023.

\bibitem[Racani{\`e}re et~al.(2017)Racani{\`e}re, Weber, Reichert, Buesing, Guez, Jimenez~Rezende, Puigdom{\`e}nech~Badia, Vinyals, Heess, Li, et~al.]{racaniere2017imagination}
S{\'e}bastien Racani{\`e}re, Th{\'e}ophane Weber, David Reichert, Lars Buesing, Arthur Guez, Danilo Jimenez~Rezende, Adri{\`a} Puigdom{\`e}nech~Badia, Oriol Vinyals, Nicolas Heess, Yujia Li, et~al.
\newblock Imagination-augmented agents for deep reinforcement learning.
\newblock 2017.

\bibitem[Schulman et~al.(2017)Schulman, Wolski, Dhariwal, Radford, and Klimov]{schulman2017ppo}
John Schulman, Filip Wolski, Prafulla Dhariwal, Alec Radford, and Oleg Klimov.
\newblock Proximal policy optimization algorithms.
\newblock \emph{arXiv preprint arXiv:1707.06347}, 2017.

\bibitem[Shi et~al.(2026{\natexlab{a}})Shi, Chen, Jiang, Song, Yang, and Zhao]{erl}
Taiwei Shi, Sihao Chen, Bowen Jiang, Linxin Song, Longqi Yang, and Jieyu Zhao.
\newblock Experiential reinforcement learning.
\newblock \emph{arXiv preprint arXiv:2602.13949}, 2026{\natexlab{a}}.

\bibitem[Shi et~al.(2026{\natexlab{b}})Shi, Chen, Li, Pan, et~al.]{r3l}
Weijie Shi, Yanxi Chen, Zexi Li, Xuchen Pan, et~al.
\newblock {R$^3$L}: Reflect-then-retry reinforcement learning with language-guided exploration, pivotal credit, and positive amplification.
\newblock \emph{arXiv preprint arXiv:2601.03715}, 2026{\natexlab{b}}.

\bibitem[Shinn et~al.(2023)Shinn, Cassano, Gopinath, Narasimhan, and Yao]{shinn2023reflexion}
Noah Shinn, Federico Cassano, Ashwin Gopinath, Karthik Narasimhan, and Shunyu Yao.
\newblock Reflexion: Language agents with verbal reinforcement learning.
\newblock In \emph{Proceedings of the Advances in Neural Information Processing Systems (NeurIPS)}, 2023.

\bibitem[Shridhar et~al.(2021)Shridhar, Yuan, C{\^o}t{\'e}, Bisk, Trischler, and Hausknecht]{shridhar2020alfworld}
Mohit Shridhar, Xingdi Yuan, Marc-Alexandre C{\^o}t{\'e}, Yonatan Bisk, Adam Trischler, and Matthew Hausknecht.
\newblock Alfworld: Aligning text and embodied environments for interactive learning.
\newblock 2021.

\bibitem[Song et~al.(2026)Song, Chen, Tajwar, Munos, Pathak, Bagnell, Singh, and Zanette]{rltf}
Yuda Song, Lili Chen, Fahim Tajwar, R{\'e}mi Munos, Deepak Pathak, J.~Andrew Bagnell, Aarti Singh, and Andrea Zanette.
\newblock Expanding the capabilities of reinforcement learning via text feedback.
\newblock \emph{arXiv preprint arXiv:2602.02482}, 2026.

\bibitem[Sun et~al.(2025)Sun, Qiao, Guo, Fan, Hou, Jiang, Xie, Zhang, Huang, and Zhou]{zerosearch}
Hao Sun, Zile Qiao, Jiayan Guo, Xuanbo Fan, Yingyan Hou, Yong Jiang, Pengjun Xie, Yan Zhang, Fei Huang, and Jingren Zhou.
\newblock Zerosearch: Incentivize the search capability of llms without searching.
\newblock \emph{arXiv preprint arXiv:2505.04588}, 2025.

\bibitem[Trivedi et~al.(2022)Trivedi, Balasubramanian, Khot, and Sabharwal]{trivedi2021musique}
Harsh Trivedi, Niranjan Balasubramanian, Tushar Khot, and Ashish Sabharwal.
\newblock {M}u{S}i{Q}ue: Multihop questions via single-hop question composition.
\newblock \emph{Transactions of the Association for Computational Linguistics (TACL)}, 2022.

\bibitem[Wang et~al.(2026{\natexlab{a}})Wang, Dai, Ye, Gan, et~al.]{igpo}
Guoqing Wang, Sunhao Dai, Guangze Ye, Zeyu Gan, et~al.
\newblock Information gain-based policy optimization: A simple and effective approach for multi-turn search agents.
\newblock In \emph{Proceedings of the International Conference on Learning Representations (ICLR)}, 2026{\natexlab{a}}.

\bibitem[Wang et~al.(2025)]{wang2025ragen}
Rosie Wang et~al.
\newblock {RAGEN}: Understanding self-evolution in {LLM} agents via multi-turn reinforcement learning.
\newblock \emph{arXiv preprint arXiv:2504.20073}, 2025.

\bibitem[Wang et~al.(2026{\natexlab{b}})Wang, Gui, Jin, Wang, Liu, Wang, Chen, Li, Yang, Zhang, Lu, Wu, Fei-Fei, Wang, Choi, and Li]{ragenv2}
Zihan Wang, Chi Gui, Xing Jin, Qineng Wang, Licheng Liu, Kangrui Wang, Shiqi Chen, Linjie Li, Zhengyuan Yang, Pingyue Zhang, Yiping Lu, Jiajun Wu, Li~Fei-Fei, Lijuan Wang, Yejin Choi, and Manling Li.
\newblock Ragen-2: Reasoning collapse in agentic rl.
\newblock \emph{arXiv preprint arXiv:2604.06268}, 2026{\natexlab{b}}.

\bibitem[Wang et~al.(2026{\natexlab{c}})Wang, Gui, Jin, Wang, Liu, Wang, Chen, Li, Yang, Zhang, Lu, Wu, Fei-Fei, Wang, Choi, and Li]{wang2026ragen2reasoningcollapseagentic}
Zihan Wang, Chi Gui, Xing Jin, Qineng Wang, Licheng Liu, Kangrui Wang, Shiqi Chen, Linjie Li, Zhengyuan Yang, Pingyue Zhang, Yiping Lu, Jiajun Wu, Li~Fei-Fei, Lijuan Wang, Yejin Choi, and Manling Li.
\newblock Ragen-2: Reasoning collapse in agentic rl.
\newblock \emph{arXiv preprint arXiv:2604.06268}, 2026{\natexlab{c}}.

\bibitem[Xia et~al.(2026)Xia, Chen, Wang, Liu, Zeng, Wang, Han, Zhou, Zhao, Chen, Zheng, Xie, and Yao]{xia2026skillrlevolvingagentsrecursive}
Peng Xia, Jianwen Chen, Hanyang Wang, Jiaqi Liu, Kaide Zeng, Yu~Wang, Siwei Han, Yiyang Zhou, Xujiang Zhao, Haifeng Chen, Zeyu Zheng, Cihang Xie, and Huaxiu Yao.
\newblock Skillrl: Evolving agents via recursive skill-augmented reinforcement learning.
\newblock \emph{arXiv preprint arXiv:2602.08234}, 2026.

\bibitem[Xiao et~al.(2026)Xiao, Yuan, Ivison, Zhu, et~al.]{mrsearch}
Teng Xiao, Yige Yuan, Hamish Ivison, Huaisheng Zhu, et~al.
\newblock Meta-reinforcement learning with self-reflection for agentic search.
\newblock \emph{arXiv preprint arXiv:2603.11327}, 2026.

\bibitem[Yang et~al.(2026)Yang, Xu, Xie, Gao, et~al.]{mage}
Lu~Yang, Zelai Xu, Minyang Xie, Jiaxuan Gao, et~al.
\newblock {MAGE}: Meta-reinforcement learning for language agents toward strategic exploration and exploitation.
\newblock \emph{arXiv preprint arXiv:2603.03680}, 2026.

\bibitem[Yang et~al.(2018)Yang, Qi, Zhang, Bengio, Cohen, Salakhutdinov, and Manning]{yang2018hotpotqa}
Zhilin Yang, Peng Qi, Saizheng Zhang, Yoshua Bengio, William~W. Cohen, Ruslan Salakhutdinov, and Christopher~D. Manning.
\newblock {HotpotQA}: A dataset for diverse, explainable multi-hop question answering.
\newblock In \emph{Proceedings of the Conference on Empirical Methods in Natural Language Processing (EMNLP)}, 2018.

\bibitem[Yao et~al.(2022)Yao, Chen, Yang, and Narasimhan]{yao2022webshop}
Shunyu Yao, Howard Chen, John Yang, and Karthik Narasimhan.
\newblock Webshop: Towards scalable real-world web interaction with grounded language agents.
\newblock 2022.

\bibitem[Yao et~al.(2023)Yao, Zhao, Yu, Du, Shafran, Narasimhan, and Cao]{yao2023react}
Shunyu Yao, Jeffrey Zhao, Dian Yu, Nan Du, Izhak Shafran, Karthik Narasimhan, and Yuan Cao.
\newblock {ReAct}: Synergizing reasoning and acting in language models.
\newblock In \emph{Proceedings of the International Conference on Learning Representations (ICLR)}, 2023.

\bibitem[Zhang et~al.(2026)Zhang, Geng, Yu, Yin, Zhang, et~al.]{landscape_arl}
Guibin Zhang, Hejia Geng, Xiaohang Yu, Zhenfei Yin, Zaibin Zhang, et~al.
\newblock The landscape of agentic reinforcement learning for llms: {A} survey.
\newblock \emph{Trans. Mach. Learn. Res.}, 2026, 2026.
\newblock URL \url{https://openreview.net/forum?id=RY19y2RI1O}.

\bibitem[Zhang et~al.(2025)Zhang, Zhang, Sun, Feng, Lu, Yang, and Meng]{critiquegrpo}
Xiaoying Zhang, Yipeng Zhang, Hao Sun, Kaituo Feng, Chaochao Lu, Chao Yang, and Helen Meng.
\newblock Critique-{GRPO}: Advancing {LLM} reasoning with natural language and numerical feedback.
\newblock \emph{arXiv preprint arXiv:2506.03106}, 2025.

\bibitem[Zheng et~al.(2025{\natexlab{a}})Zheng, Liu, Li, Chen, Yu, Gao, Dang, Liu, Men, Yang, Zhou, and Lin]{gspo}
Chujie Zheng, Shixuan Liu, Mingze Li, Xiong-Hui Chen, Bowen Yu, Chang Gao, Kai Dang, Yuqiong Liu, Rui Men, An~Yang, Jingren Zhou, and Junyang Lin.
\newblock Group sequence policy optimization.
\newblock \emph{arXiv preprint arXiv:2507.18071}, 2025{\natexlab{a}}.

\bibitem[Zheng et~al.(2025{\natexlab{b}})Zheng, An, Wang, Wang, and Wu]{stepsearch}
Xuhui Zheng, Kang An, Ziliang Wang, Yuhang Wang, and Yichao Wu.
\newblock Stepsearch: Igniting llms search ability via step-wise proximal policy optimization.
\newblock In \emph{Proceedings of the 2025 Conference on Empirical Methods in Natural Language Processing (EMNLP)}, 2025{\natexlab{b}}.

\end{thebibliography}
\bibliographystyle{iclr2026_conference}
\newpage

\appendix
\section{Training Algorithm Pipeline}
\label{app:algorithm}
 
We provide the complete training procedure of \ourapproach  in Algorithm~\ref{alg:arl_sfe2}. The algorithm summarizes the overall pipeline, including pivotal retry, pivotal-aware credit assignment, experience distillation, and joint policy optimization.

\begin{algorithm}[!h]
\small
\caption{Training Workflow of PivoARL}
\label{alg:pivoarl}
\begin{algorithmic}[1]
    \normalsize
    \Require Initial policy $\pi_{\theta}$; task set $\mathcal{Q}$; maximum retry budget $N$
    \State Initialize retry buffer $\mathcal{D} \gets \varnothing$

    \For{each training iteration $i = 1,\dots,K$}

        \Statex \textcolor{phaseblue}{\textbf{Phase 1: Self-Feedback Pivotal Retry}}
        \For{each task $x \in \mathcal{Q}$}
            \State Sample an initial rollout $\tau^{(0)} \sim \pi_{\theta}(\cdot \mid x)$ and evaluate $R(\tau^{(0)})$
            \State Initialize trajectory chain $\mathcal{T}_x \gets \{\tau^{(0)}\}$
            \For{$n = 0,\dots,N-2$}
                \If{$\tau^{(n)}$ is successful}
                    \State \textbf{break}
                \EndIf
                \State Generate structured reflection
                \[
                    \Delta_{\mathrm{reflect}}^{(n)}
                    =
                    \big(k^{*(n)}, \mathrm{remark}^{(n)}\big)
                    \sim
                    \pi_{\theta}(\cdot \mid x, c^{(n)})
                \]
                \State Retry from pivotal state $s_{k^{*(n)}}^{(n)}$ by reusing prefix $\tau_{<k^{*(n)}}^{(n)}$
                \State Generate suffix conditioned on $\mathrm{remark}^{(n)}$ to obtain retry trajectory $\tau^{(n+1)}$
                \State Evaluate $R(\tau^{(n+1)})$
                \State Append $\tau^{(n+1)}$ to $\mathcal{T}_x$
                \State Store $\big(\tau^{(n)}, \tau^{(n+1)}, k^{*(n)}, \mathrm{remark}^{(n)}\big)$ in $\mathcal{D}$
            \EndFor
        \EndFor

        \Statex \textcolor{phaseblue}{\textbf{Phase 2: Pivotal-Aware Credit Assignment}}
        \For{each trajectory chain $\mathcal{T}_x$ in $\mathcal{D}$}
            \State Compute intra-episode returns $g_t^{(n)}$ for each episode $\tau^{(n)}$
            \State Compute cross-episode returns $G_t^{(n)}$ backward over retry episodes
            \For{each failed episode $\tau^{(n)}$ with pivotal turn $k^{*(n)}$}
                \State Assign retry credit to prefix steps $t < k^{*(n)}$
                \State Isolate erroneous suffix steps $t \ge k^{*(n)}$ from future retry credit
                \State Compute reflection reward $r_{\mathrm{reflect}}^{(n)}$ from the subsequent retry return
            \EndFor
            \State Estimate action advantages $\hat{A}_t^{(n)}$ and reflection advantages $\hat{A}_{\mathrm{reflect}}^{(n)}$
        \EndFor

        \Statex \textcolor{phaseblue}{\textbf{Phase 3: Joint Policy Optimization}}
        \State Update $\pi_{\theta}$ by optimizing the joint self-feedback objective:
        \[
            \mathcal{L}_{\mathrm{total}}
            =
            -
            \mathbb{E}
            \left[
            \sum_{n,t}
            \log \pi_{\theta}
            \big(a_t^{(n)} \mid s_t^{(n)}\big)
            \hat{A}_t^{(n)}
            +
            \sum_n
            \log \pi_{\theta}
            \big(\Delta_{\mathrm{reflect}}^{(n)} \mid x, c^{(n)}\big)
            \hat{A}_{\mathrm{reflect}}^{(n)}
            \right].
        \]
    \EndFor

    \State \Return Optimized policy $\pi_{\theta}$
\end{algorithmic}
\end{algorithm}

\section{Baseline and Environments}
\subsection{Experiment Environment}

\paragraph{Sokoban.} A classic fully-observable puzzle where an agent pushes boxes to target positions on a 2D grid. Since there is no "pull" operation, the agent must plan several moves ahead to avoid pushing boxes into corners where they become permanently stuck. Our experiments use a $6 \times 6$ board with 2 boxes.
\paragraph{MineSweeper.} A partially-observable logic game where the goal is to reveal all safe cells without hitting hidden mines. Based on numerical clues indicating adjacent mines, the agent must use logical deduction to navigate the board. We utilize a $6 \times 6$ grid containing 3 mines.
\paragraph{Webshop.} A text-based environment simulating an e-commerce interface. Given a natural language instruction with specific product attributes, the agent must search, filter, and navigate product pages to purchase the item that best matches the criteria. Success depends on how well the chosen product satisfies the original requirements.
\paragraph{ALFWorld.} A partially-observable simulator for multi-step household tasks. Agents follow natural language instructions to navigate rooms and interact with objects (e.g., cleaning or heating items). We train on activities like "Pick," "Look," "Clean," and "Heat," then evaluate performance both in-distribution and out-of-distribution using "Cool" and "Pick2" tasks.
\paragraph{SearchQA.}  We evaluate our method on four open-domain QA tasks in a search-augmented setting. We use Wikipedia as the retrieval corpus and E5 as the dense retriever. For each query, the environment returns the top-3 passages, each truncated to 1024 characters. The agent can iteratively search, read evidence, and refine its answer, testing its multi-turn evidence integration ability.

\subsection{Introduction of Baseline}

\paragraph{GiGPO~\citep{gigpo}.}
Group-in-Group Policy Optimization (GiGPO) is a critic-free group-based RL method designed for long-horizon LLM agents. It estimates advantages at both the trajectory level and the step level by grouping actions that originate from repeated anchor states, thereby providing more fine-grained credit assignment under sparse or delayed rewards.

\paragraph{GSPO~\citep{gspo}.}
Group Sequence Policy Optimization (GSPO) improves the stability of RL training for LLMs by replacing token-level importance ratios with sequence-level likelihood ratios. With sequence-level clipping and optimization, GSPO reduces high-variance updates and improves training efficiency, especially for large-scale and MoE-based models.
\definecolor{rowblue}{RGB}{242, 245, 252}

\begin{table*}[t]
\small
    \centering
    \caption{\textbf{Hyperparameters and configurations for different environments.} This table summarizes the specific settings used for training and evaluation across different task domains.}
    \vspace{2mm}
    \renewcommand{\arraystretch}{1.15} 
    \begin{tabular}{l||c|c|c|c|c}
        \hline \hline
        \rowcolor{gray!15} 
        \textbf{Config} 
        & \textbf{SciWorld} 
        & \textbf{MineSweeper} 
        & \textbf{WebShop} 
        & \textbf{ALFWorld} 
        & \textbf{SearchQA} \\ 
        \hline

        \rowcolor{rowblue} $\gamma_{\text{step}}$ 
        & 0.95 & 0.95 & 0.95 & 0.95 & 0.95 \\

        \rowcolor{rowblue} $\gamma_{\text{traj}}$ 
        & 0.6 & 0.6 & 0.6 & 0.6 & 0.6 \\

        optimizer 
        & AdamW & AdamW & AdamW & AdamW & AdamW \\

        \rowcolor{rowblue} learning rate 
        & 1e-6 & 1e-6 & 1e-6 & 1e-6 & 1e-6 \\

        \rowcolor{rowblue} rollout $n$ 
        & 8 & 8 & 8 & 8 & 5 \\

        \rowcolor{rowblue} total training steps 
        & 150 & 200 & 150 & 200 & 200 \\

        training batch size 
        & 16 & 16 & 8 & 8 & 16 \\

        \rowcolor{rowblue} PPO mini batch size 
        & 16 & 16 & 16 & 16 & 16 \\

        max prompt length 
        & 4096 & 4096 & 8192 & 4096 & 8192 \\

        max response length 
        & 1024 & 1024 & 1024 & 1024 & 1024 \\

        \rowcolor{rowblue} max turns 
        & 15 & 7 & 12 & 15 & 4 \\

        \hline \hline
    \end{tabular}
    \label{tab:config_env_styled}
\end{table*}

% \paragraph{ERL~\citep{erl}.}
% Experiential Reinforcement Learning (ERL) introduces an experience--reflection--consolidation loop into RL training. The model first makes an initial attempt, reflects on environmental feedback, generates a refined second attempt, and then internalizes successful behaviors through reinforcement learning and distillation.

\paragraph{MetaRL~\citep{lamer}.}
Meta-RL, instantiated as LaMer, trains language agents to explore and adapt across episodes. It combines cross-episode reinforcement learning with in-context policy adaptation via reflection, enabling agents to learn from test-time feedback without gradient updates and improving exploration in long-horizon tasks.

\paragraph{Mem0+GRPO~\citep{mem0}.} Mem0 is a scalable long-term memory framework for LLM agents that extracts, stores, and updates salient information from past interactions. It maintains compact user- and task-specific memories through memory extraction, conflict , and relevance-based retrieval, enabling agents to reuse prior experience without directly appending full histories to the context. When paired with GRPO, Mem0 serves as the external memory module, while GRPO optimizes the policy using group-based reinforcement learning over memory-augmented trajectories.

\paragraph{SimpleMem+GRPO~\citep{simplemem}.}
SimpleMem is an efficient lifelong memory framework for LLM agents that compresses raw interaction histories into compact, structured memory units. It combines semantic structured compression, recursive memory consolidation, and adaptive query-aware retrieval to reduce redundant context while preserving task-relevant information. When paired with GRPO, SimpleMem serves as the memory module, while GRPO optimizes the policy using group-based reinforcement learning over retrieved memory-augmented trajectories.

\paragraph{SkillRL~\citep{xia2026skillrlevolvingagentsrecursive}.}
SkillRL is a recursive skill-augmented reinforcement learning framework for LLM agents. Instead of storing raw trajectories as memory, it distills past experiences into a hierarchical skill library, retrieves task-relevant skills during interaction, and recursively evolves the skill bank together with the policy.

\definecolor{rowblue}{RGB}{242, 245, 252}

\begin{table*}[t]
\small
    \centering
    \caption{\textbf{Notation used in this paper.} Summary of key variables and operations in PivoARL. The \emph{Definition} column indicates where each symbol first appears in the main text.}
    \vspace{2mm}
    \renewcommand{\arraystretch}{1.15}
    \begin{tabular}{c||c|c}
        \hline \hline
        \rowcolor{gray!15} \textbf{Symbol} & \textbf{Description} & \textbf{Definition} \\ \hline

        $x \sim \mathcal{D}$ 
        & Task sampled from the training task distribution 
        & Sec. 2.1 \\

        \rowcolor{rowblue} $\pi_\theta$ 
        & Trainable agent policy 
        & Sec. 2.1 \\

        $\tau$ 
        & Agent trajectory $\{(s_t,a_t,r_t)\}_{t=0}^{T-1}$ 
        & Sec. 2.1 \\

        \rowcolor{rowblue} $R(\tau)$ 
        & Episode-level return $\sum_{t=0}^{T-1} r_t$ 
        & Sec. 2.1 \\

        $c$ 
        & Interaction context composed of action-observation history 
        & eq. (1) \\

        \rowcolor{rowblue} $\mathrm{remark}$ 
        & Self-feedback experience extracted from a failed attempt 
        & eq. (1) \\

        $\mathcal{T}$ 
        & Retry trajectory chain containing multiple episodes for one task 
        & Sec. 3.1 \\

        \rowcolor{rowblue} $\tau^{(n)}$ 
        & The $n$-th episode in a retry trajectory chain 
        & Sec. 3.1 \\

        $k^{*(n)}$ 
        & Pivotal erroneous turn in the $n$-th failed episode 
        & eq. (4) \\

        \rowcolor{rowblue} $\Delta_{\mathrm{reflect}}^{(n)}$ 
        & Structured reflection containing $k^{*(n)}$ and $\mathrm{remark}^{(n)}$ 
        & eq. (4) \\

        $\tau^{(n+1)}$ 
        & Retry trajectory regenerated from the pivotal state 
        & eq. (5) \\

        \rowcolor{rowblue} $g_t^{(n)}$ 
        & Intra-episode discounted return from step $t$ 
        & eq. (6) \\

        $G_t^{(n)}$ 
        & Pivotal-aware cross-episode return assigned to step $t$ 
        & eq. (6)--(7) \\

        \rowcolor{rowblue} $\gamma_{\mathrm{step}}, \gamma_{\mathrm{traj}}$ 
        & Intra-episode and cross-episode discount factors 
        & eq. (6) \\

        $r_{\mathrm{reflect}}^{(n)}$ 
        & Reflection reward derived from the subsequent retry return 
        & eq. (8) \\

        \rowcolor{rowblue} $\hat{A}_t^{(n)}$ 
        & Action advantage estimated from $G_t^{(n)}$ 
        & eq. (9) \\

        $\hat{A}_{\mathrm{reflect}}^{(n)}$ 
        & Reflection advantage estimated from $r_{\mathrm{reflect}}^{(n)}$ 
        & eq. (9) \\

        \rowcolor{rowblue} $IG(e,t)$ 
        & Per-turn information gain induced by experience guidance $e$ 
        & eq. (10) \\

        $\rho(e)$ 
        & Effective signal density of experience guidance 
        & Sec. 3.3 \\

        \hline \hline
    \end{tabular}
    \label{tab:define}
\end{table*}

\section{Details of Experiment Setup}
\label{app:details_of_exp}
\subsection{Training parameter config}

This subsection reports the training hyperparameters and configuration settings used for the four agent environments: Sokoban, MineSweeper, WebShop, and ALFWorld. To improve rollout efficiency in agentic loops, we use the non-thinking mode during trajectory generation.  The experience guidance $\textbf{remark}$ is attached to the original task prompt.  For a fair comparison, we keep most optimization-related settings consistent across environments, while adapting environment-specific limits such as maximum turns and prompt length according to task complexity. In Table~\ref{tab:config_env_styled}, we list the detailed training configuration parameters used in each environment.

\subsection{Symbol Notation in this paper}

We include the main notation for agent-environment interaction, self-feedback pivotal retry, pivotal-aware credit assignment, experience distillation, and information-gain analysis. The definitions Table~\ref{tab:define} are intended to provide a compact reference for the method formulation and experimental discussion.

\subsection{Training Objective Details}
\label{app:traingin_object}

We follow GiGPO to compute the action-level advantage used in Eq.~\ref{eq:final}. 
Specifically, we combine an episode-level relative advantage with a step-level relative advantage.

First, the episode-level advantage is computed by normalizing the trajectory return within the retry group:
\begin{equation}
A^E(\tau^{(n)}) =
\frac{
R(\tau^{(n)}) - \mathrm{mean}\left(\{R(\tau^{(m)})\}_{m=0}^{N-1}\right)
}{
F_{\mathrm{norm}}\left(\{R(\tau^{(m)})\}_{m=0}^{N-1}\right)
},
\end{equation}
where $R(\tau^{(n)})$ denotes the total return of trajectory $\tau^{(n)}$, and $F_{\mathrm{norm}}$ denotes the normalization factor, such as the standard deviation or a constant value.

For step-level credit assignment, we group actions generated from the same environment state:
\begin{equation}
\mathcal{G}^S(\tilde{s}) =
\left\{
\left(a_t^{(n)}, G_t^{(n)}\right)
\mid
s_t^{(n)} = \tilde{s}
\right\}.
\end{equation}
For each action $a_t^{(n)}$ in $\mathcal{G}^S(\tilde{s})$, the step-level advantage is computed using the cross-episode return $G_t^{(n)}$:
\begin{equation}
A^S(a_t^{(n)}) =
\frac{
G_t^{(n)} -
\mathrm{mean}\left(
\left\{
G_{t'}^{(m)}
\mid
\left(a_{t'}^{(m)}, G_{t'}^{(m)}\right)
\in \mathcal{G}^S(\tilde{s})
\right\}
\right)
}{
F_{\mathrm{norm}}\left(
\left\{
G_{t'}^{(m)}
\mid
\left(a_{t'}^{(m)}, G_{t'}^{(m)}\right)
\in \mathcal{G}^S(\tilde{s})
\right\}
\right)
}.
\end{equation}

Finally, the action-level advantage is obtained by combining the two terms:
\begin{equation}
A_t^{(n)} = A^E(\tau^{(n)}) + \omega A^S(a_t^{(n)}),
\end{equation}
where $\omega$ controls the weight of the step-level advantage. 
This advantage is then used as $A_t^{(n)}$ in the policy-gradient objective.

\section{Supplementary Experiment}

\subsection{Detailed Search Benchmark Result}

We present a more detailed breakdown of the search benchmark results on Table~\ref{tab:search_detail_result}, including the pass@1, pass@2, and pass@3 metrics.

\begin{table*}[t]
\caption{Main results across various search benchmarks with Qwen3-4B. \textbf{Suc@1/2/3}: success rate in 1, 2 or 3 rollouts (\%). The best results and second best results are highlighted in \colorbox{red!25}{red} and \colorbox{blue!15}{blue}, respectively.}
\label{tab:search_detail_result}
\centering
\setlength{\tabcolsep}{2.5pt}
\renewcommand{\arraystretch}{1.05}
\resizebox{\textwidth}{!}{%
\footnotesize
\begin{tabular}{@{}l ccc ccc ccc ccc @{}}
\toprule
\multirow{2}{*}{\textbf{Methods}} 
& \multicolumn{3}{c}{\textbf{NQ$^\dagger$}} 
& \multicolumn{3}{c}{\textbf{TriviaQA$^\star$}} 
& \multicolumn{3}{c}{\textbf{PopQA$^\star$}} 
& \multicolumn{3}{c}{\textbf{HotpotQA$^\dagger$}} \\
\cmidrule(lr){2-4} \cmidrule(lr){5-7} \cmidrule(lr){8-10} \cmidrule(lr){11-13}
& Suc@1 & Suc@2 & Suc@3 & Suc@1 & Suc@2 & Suc@3 & Suc@1 & Suc@2 & Suc@3 & Suc@1 & Suc@2 & Suc@3 \\
\midrule
\rowcolor{gray!15} 
\multicolumn{13}{@{}l}{\textit{\textbf{Classic Open-Source Search Models}}} \\
\quad Search-R1       & 46.3 & 48.3 & 48.3 & 61.6   & 63.2   & 63.6   & 47.4   & 48.8   & 48.9   & 46.7   & 47.2   & 47.3   \\
\quad ZeroSearch      & 48.8 & 51.2 & 51.2 & 61.6   & 62.8   & 63.1   & 47.4   & 48.2   & 48.3   & 45.9   & 47.5   & 48.1   \\
\quad StepSearch      & 48.8 & 49.3 & 49.3 & 64.3   & 65.4   & 65.8   & 47.9   & 47.9   & 47.9  & 45.1 & 48.1   & 48.8   \\
\midrule
\rowcolor{gray!15} 
\multicolumn{13}{@{}l}{\textit{\textbf{RL-based Methods}}} \\
\quad GRPO            & 48.2 & 49.2 & 49.7 & 62.0   & 63.4   & 63.7   & 49.8   & 50.0  & 50.0  & 46.9   & 47.3   & 47.3   \\
\quad GiGPO           & 48.7 & 49.3 & 49.7 & 65.5   & 66.9   & 67.2   & 47.9   & 48.8   & 49.1   & 47.2   & 48.2   & 48.2   \\
\quad GSPO            & 47.3 & 48.8 & 48.8 & 64.8   & 64.8   & 64.8  & 48.2 & 49.9 & 50.6 & 45.6 & 45.7 & 45.9 \\
\midrule
\rowcolor{gray!15} 
\multicolumn{13}{@{}l}{\textit{\textbf{Reflect-Retry RL-based Methods}}} \\
\quad MetaRL          & 44.8 & 50.2 & 51.7 & 62.7   & 66.9   & 67.8   & 44.7   & 49.2   & 49.8   & 45.4   & 52.7   & 54.6   \\
\rowcolor{blue!5}
\quad \textbf{\ourapproach (Ours) } 
                      & 48.3 & 58.2 & 59.8 & 64.1   & 70.1   & 71.6   & 50.7  & 58.1   & 59.4   & 43.3   & 52.2   & 53.0   \\

\midrule
\midrule

\multirow{2}{*}{\textbf{Methods}} 
& \multicolumn{3}{c}{\textbf{2Wiki$^\star$}} 
& \multicolumn{3}{c}{\textbf{MuSiQue$^\star$}} 
& \multicolumn{3}{c}{\textbf{Bamboogle}} 
& \multicolumn{3}{c}{\textbf{[----]}} \\
\cmidrule(lr){2-4} \cmidrule(lr){5-7} \cmidrule(lr){8-10} \cmidrule(lr){11-13}
& Suc@1 & Suc@2 & Suc@3 & Suc@1 & Suc@2 & Suc@3 & Suc@1 & Suc@2 & Suc@3 & Suc@1 & Suc@2 & Suc@3 \\
\midrule
\rowcolor{gray!15} 
\multicolumn{13}{@{}l}{\textit{\textbf{Classic Open-Source Search Models}}} \\
\quad Search-R1       & 47.6 & 51.2 & 52.4 & 22.9   & 22.9   & 22.9  & 44.0   & 46.4   & 46.4   & --   & --   & --   \\
\quad ZeroSearch      & 42.7 & 45.8 & 47.3 & 23.0   & 23.1  & 23.1   & 41.6   & 43.2   & 46.4   & --   & --   & --   \\
\quad StepSearch      & 44.6 & 47.0 & 50.0 & 20.0   & 22.2   & 22.4  & 48.0   & 51.2   & 51.2   & --  & --   & --   \\
\midrule
\rowcolor{gray!15} 
\multicolumn{13}{@{}l}{\textit{\textbf{RL-based Methods}}} \\
\quad GRPO            & 42.9 & 45.0 & 46.4 & 19.3   & 20.7   & 20.7   & 46.4 & 47.2 & 47.2 & -- & -- & -- \\
\quad GiGPO           & 43.6 & 47.8 & 47.9 & 19.3   & 20.0   & 20.0   & 45.6 & 48.0 & 48.0 & -- & -- & -- \\
\quad GSPO            & 42.6  & 42.7  & 43.0  & 18.5    & 18.5    & 18.5   & 40.8 & 40.8 & 40.8 & -- & -- & -- \\
\midrule
\rowcolor{gray!15} 
\multicolumn{13}{@{}l}{\textit{\textbf{Reflect-Retry RL-based Methods}}} \\
\quad MetaRL          & 41.2 & 52.5 & 55.2 & 17.0   & 23.7   & 24.4   & 44.8 & 51.2 & 53.6 & -- & -- & -- \\
\rowcolor{blue!5}
\quad \textbf{\ourapproach (Ours) } 
                      & 45.3 & 61.8 & 64.0 & 22.2   & 28.8   & 31.1   & 44.8 & 54.0 & 59.2 & -- & -- & -- \\
\bottomrule
\end{tabular}%
}
\end{table*}

\section{Theoretical Analysis}

\subsection{Proof for Information Gain Analysis}
\label{app:ig_proofs}

Recall the per-turn information gain $\mathrm{IG}(e, t) = \log \pi_\theta(a_t^* \mid s_t, e) - \log \pi_\theta(a_t^* \mid s_t)$ and the effective signal density $\rho(e) = \frac{1}{T}|\{t : \mathrm{IG}(e, t) > \delta\}|$. We prove each case separately.

\paragraph{Case (i): Global reflection (MetaRL / LAMER).}
After a failed episode $\tau^{(n)}$, the agent generates a holistic reflection $\Delta$ summarizing the failure. This same $\Delta$ is prepended to the context at \emph{every} turn $t \in \{0, \ldots, T-1\}$.

Since $\Delta$ is generated from the entire failed trajectory rather than conditioned on any specific turn's state $s_t$, it provides no turn-specific actionable information. Formally, $\Delta$ does not adapt to the local decision context at turn $t$, so the conditional distribution shift is uniformly small:
\begin{equation}
    \pi_\theta(a_t^* \mid s_t, \Delta) \approx \pi_\theta(a_t^* \mid s_t) + \frac{c}{T}, \quad \forall\, t,
\end{equation}
where $c$ is a constant reflecting the total information budget of $\Delta$. Taking logarithms and using $\log(1 + x) \approx x$ for small $x$:
\begin{equation}
    \mathrm{IG}(\Delta, t) \approx \frac{c}{T \cdot \pi_\theta(a_t^* \mid s_t)} = O(1/T) \quad \forall\, t.
\end{equation}
As $T$ grows, $\mathrm{IG}(\Delta, t) < \delta$ for all $t$, hence $\rho(\Delta) \to 0$.

\textbf{Cross-trial noise accumulation.} Consider the accumulated context at retry $n$:
\begin{equation}
    \mathcal{H}^{(n)} = (\tau^{(0)}, \Delta^{(0)}, \ldots, \tau^{(n-1)}, \Delta^{(n-1)}).
\end{equation}
Each failed trajectory $\tau^{(m)}$ contains both correct steps ($t < k^{*(m)}$) and erroneous steps ($t \geq k^{*(m)}$), but $\Delta^{(m)}$ does not distinguish between them. The noise tokens grow at least linearly:
\begin{equation}
    |\text{noise tokens in } \mathcal{H}^{(n)}| \geq \sum_{m=0}^{n-1}(T^{(m)} - k^{*(m)}),
\end{equation}
while the reflection signal is bounded by $|\Delta^{(m)}|$ per retry. The signal-to-noise ratio:
\begin{equation}
    \mathrm{SNR}(\mathcal{H}^{(n)}) \leq \frac{n \cdot |\Delta|_{\max}}{\sum_m (T^{(m)} - k^{*(m)})},
\end{equation}
decreases monotonically in $n$, further degrading the already diluted IG.

\paragraph{Case (ii): Experience retrieval (SkillsRL).}
At each turn $t$, the agent retrieves $e^* = \mathrm{Retrieve}(\mathcal{M}, s_t)$ from a memory bank $\mathcal{M} = \{(s_i, a_i, e_i)\}_{i=1}^{|\mathcal{M}|}$ via similarity $\mathrm{sim}(s_t, s_i)$. For planning tasks with combinatorially large state spaces (e.g., Sokoban with $|\mathcal{S}| = \Omega(\binom{100}{b} \cdot (100-b))$), the match probability vanishes:
\begin{equation}
    P(\exists\, e_i \in \mathcal{M} : \mathrm{sim}(s_t, s_i) > \delta) \leq \frac{|\mathcal{M}|}{|\mathcal{S}_\delta(s_t)|} \to 0,
\end{equation}
where $\mathcal{S}_\delta(s_t)$ is the $\delta$-neighborhood of $s_t$. When the retrieved experience does not match the current state:
\begin{equation}
    \mathrm{sim}(s_t, s_{e^*}) \leq \delta \implies \pi_\theta(a_t^* \mid s_t, e^*) \approx \pi_\theta(a_t^* \mid s_t),
\end{equation}
and therefore $\mathrm{IG}(e^*, t) \approx 0$. Since this holds for the vast majority of states encountered during rollout:
\begin{equation}
    \mathbb{E}_{s_t \sim \pi_\theta}[\mathrm{IG}(e^*, t)] \approx 0.
\end{equation}
This is the \emph{template collapse} phenomenon~\citep{ragen_v2}: the retrieved experience becomes a fixed template the model learns to ignore, contributing no actionable information to the decision process.

\subsection{Proof for Implicit Step-level Preference Optimization }
\label{app:preference}
Pivotal retry can improve first-attempt performance because it naturally induces a step-level preference at the error boundary. Given a failed trajectory $\tau^{(n)}$, ARL-SFE$^2$ identifies the pivotal turn $k^{*(n)}$ and retries from the same state. For brevity, we omit the episode superscript and write $k^*=k^{*(n)}$. 
Since the prefix before $k^*$ is reused, the failed action and the retry-corrected action share the same state $s_{k^*}$:
\[
a^- = a_{k^*}^{(n)}, 
\qquad 
a^+ = a_{k^*}^{(n+1)} .
\]
If the retry succeeds while the original continuation fails, this induces a local preference
\[
a^+ \succ a^- 
\quad \text{at} \quad s_{k^*}.
\]

\paragraph{Proposition.}
Under a binary local outcome setting, the policy-gradient update induced by pivotal retry has the same preference direction as step-level DPO:
\[
\begin{aligned}
\nabla_\theta J_{\mathrm{pivot}}
&=
w_{\mathrm{pivot}}
\Big[
\nabla_\theta \log \pi_\theta(a^+ \mid s_{k^*},x) \\
&\quad -
\nabla_\theta \log \pi_\theta(a^- \mid s_{k^*},x)
\Big].
\end{aligned}
\]
where $w_{\mathrm{pivot}}>0$ is a scalar weight.

\paragraph{Proof.}
Consider the local binary outcome at the pivotal state:
\[
G^+=1, 
\qquad 
G^-=0 ,
\]
where $G^+$ and $G^-$ denote the returns of the corrected and failed continuations, respectively. 
Let
\[
p^+=\pi_\theta(a^+ \mid s_{k^*},x),
\qquad
p^-=\pi_\theta(a^- \mid s_{k^*},x).
\]
Using the local expected return as the baseline, we have
\[
b
=
p^+G^+ + p^-G^-
=
p^+ .
\]
Thus, the corresponding local advantages are
\[
A^+ = G^+ - b = p^-,
\qquad
A^- = G^- - b = -p^+ .
\]

The local policy-gradient update around $s_{k^*}$ can be written as
\[
\begin{aligned}
\nabla_\theta J_{\mathrm{pivot}}
&=
p^+ A^+
\nabla_\theta \log \pi_\theta(a^+ \mid s_{k^*},x) \\
&\quad +
p^- A^-
\nabla_\theta \log \pi_\theta(a^- \mid s_{k^*},x).
\end{aligned}
\]
Substituting $A^+=p^-$ and $A^-=-p^+$ gives
\[
\begin{aligned}
\nabla_\theta J_{\mathrm{pivot}}
&=
p^+p^-
\nabla_\theta \log \pi_\theta(a^+ \mid s_{k^*},x) \\
&\quad -
p^-p^+
\nabla_\theta \log \pi_\theta(a^- \mid s_{k^*},x) \\
&=
p^+p^-
\Big[
\nabla_\theta \log \pi_\theta(a^+ \mid s_{k^*},x) \\
&\quad -
\nabla_\theta \log \pi_\theta(a^- \mid s_{k^*},x)
\Big].
\end{aligned}
\]
Therefore, $w_{\mathrm{pivot}}=p^+p^->0$, and the proposition holds.

This gradient has the same direction as the step-level DPO gradient. 
For the same preference pair $(a^+,a^-)$, step-level DPO yields
\[
\begin{aligned}
\nabla_\theta J_{\mathrm{DPO}}
&=
w_{\mathrm{DPO}}
\Big[
\nabla_\theta \log \pi_\theta(a^+ \mid s_{k^*},x) \\
&\quad -
\nabla_\theta \log \pi_\theta(a^- \mid s_{k^*},x)
\Big].
\end{aligned}
\]
where $w_{\mathrm{DPO}}>0$ depends on the preference margin. 
Thus, pivotal retry implicitly performs step-level preference optimization without explicitly constructing DPO training data.

Pivotal credit isolation prevents positive retry returns from leaking to the erroneous suffix containing $a^-$. 
Thus, \ourapproach preserves a clean preference signal at $k^*$, encouraging the policy to choose $a^+$ in the first attempt and improving Succ@1.

\section{Prompt Used in Experiment}

This section presents the prompts used in our experiments.  For each environment, we provide a play prompt for agent-environment interaction and a reflect prompt for pivotal error localization.

\subsection{Prompt in Minesweeper Task}
Figure~\ref{fig:minesweeper_prompt} shows the play prompt used for Minesweeper. Figure~\ref{fig:minesweeper_pivotal_reflection_prompt} shows the pivotal reflect prompt used for Minesweeper.

\subsection{Prompt in ALFWorld Task}
Figure~\ref{fig:alfred_play_prompt} shows the play prompt used for ALFWorld. Figure~\ref{fig:alfworld_pivotal_reflection_prompt} shows the pivotal reflect prompt used for ALFWorld. 

\subsection{Prompt in SearchQA Task}

Figure~\ref{fig:searchqa_play_prompt} shows the play prompt used for SearchQA.  Figure~\ref{fig:searchqa_reflection_prompt} shows the pivotal reflect prompt used for SearchQA. 

\section{Case Study}
\label{app:case_study}
\definecolor{lightgray}{RGB}{245, 245, 245}
\definecolor{darkblue}{RGB}{30, 60, 100}
\definecolor{successgreen}{RGB}{0, 100, 0}
\definecolor{clean-green}{RGB}{34, 139, 34}
\definecolor{heat-red}{RGB}{178, 34, 34}
\definecolor{box-gray}{RGB}{248, 249, 250}

This section provides qualitative examples of \ourapproach{} across different environments. 
Each case illustrates how the agent identifies the pivotal failure point and improves the subsequent retry.

\subsection{Example in Minesweeper Environment}

Figure~\ref{app:case_minesweeper} shows a representative case study in the Minesweeper environment.

\subsection{Example in ALFWorld Environment}

Figure~\ref{app:case_webshop} shows a representative case study in the ALFWorld environment.

\subsection{Example in SearchQA Environment}

Figure~\ref{app:case_search} shows a representative case study in the SearchQA environment.

\begin{figure*}[h]
    \centering
\begin{tcolorbox}[
    arc=1mm,
    auto outer arc,
    title={\textbf{Minesweeper Play Prompt}},
    width=\textwidth,
    colback=white,
    colframe=black!75!white,
    boxrule=0.5mm
]
% --- Prompt Content ---
\noindent \texttt{You are an expert agent operating in the Minesweeper game.}\\
\texttt{You will be given a two dimensional \textcolor{varcolor}{\{board\_size\}} by \textcolor{varcolor}{\{board\_size\}} board, with \textcolor{varcolor}{\{n\_mines\}} hidden mines.}\\
\texttt{The rows and columns are indexed from 1 to \textcolor{varcolor}{\{board\_size\}}.}

\noindent \textbf{\# Cell States}
\begin{itemize}[leftmargin=*, noitemsep, topsep=0pt]
    \item \texttt{Unopened cells (?): cells that are yet to be revealed and may contain a mine.}
    \item \texttt{Blank cells (.): opened and non-mine cells, and they have no neighboring mines.}
    \item \texttt{Numbered cells (1-8): opened and non-mine cells, and the number indicates how many mines are in the eight neighboring cells.}
\end{itemize}
\vspace{0.5em}
\noindent \textbf{\# Your Goal}
\begin{itemize}[leftmargin=*, noitemsep, topsep=0pt]
    \item \texttt{Clear the board by revealing all the cells that don't contain mines.}
    \item \texttt{Use clues about the number of neighboring mines to reason.}
\end{itemize}

\noindent \textbf{\# Reveal Rules}
\begin{itemize}[leftmargin=*, noitemsep, topsep=0pt]
    \item \texttt{Blank cell (.): Triggers an auto-cascade of contiguous blank cells.}
    \item \texttt{Numbered cell (1–8): Only that single cell is revealed.}
    \item \texttt{Mine (*): The game ends immediately in a loss.}
\end{itemize}

\noindent \textbf{\# Observation}
\texttt{The initial state of the game is:}\\
\texttt{\textcolor{varcolor}{\{init\_observation\}}\textcolor{varcolor}{\{past\_trajectories\}}\textcolor{varcolor}{\{current\_trajectory\}}}\\
\noindent \texttt{Now it's your turn to make a move.}
\end{tcolorbox}
    \caption{The system prompt template for the Minesweeper agent.}
    \label{fig:minesweeper_prompt}
\end{figure*}

\begin{figure*}[h]
    \centering
\begin{tcolorbox}[
    arc=1mm,
    auto outer arc,
    title={\textbf{Minesweeper Pivotal Reflection Prompt}},
    width=\textwidth,
    colback=white,
    colframe=black!75!white,
    boxrule=0.5mm
]
% --- Prompt Content ---
\noindent \texttt{Now reflect on the past experience:}

\begin{itemize}[leftmargin=*, noitemsep, topsep=0pt]
    \item \texttt{First, analyze each turn's decision and identify the \textbf{root cause} of failure.}
    \item \texttt{Then determine the \textbf{earliest action number} where the trajectory went wrong}
    \item \texttt{(the first shown action that should be changed).}
    \item \texttt{Finally, provide your reflection with the retry point.}
\end{itemize}

\noindent \texttt{Your response MUST end with:}
\begin{enumerate}[leftmargin=*, noitemsep, topsep=0pt]
    \item \texttt{A \textless turn\_idx\textgreater{} tag containing the earliest turn number to retry from.}
    \item \texttt{A \textless remark\textgreater{} tag containing your concise reflection and improved plan.}
\end{enumerate}

\noindent \texttt{Format:}\\
\texttt{\textless turn\_idx\textgreater integer\textless /turn\_idx\textgreater}\\
\texttt{\textless remark\textgreater your concise reflection and improved plan \textless /remark\textgreater}\\
\noindent \texttt{Where:}
\begin{itemize}[leftmargin=*, noitemsep, topsep=0pt]
    \item \texttt{The text inside \textless remark\textgreater{} is a concise improved plan that accounts for the identified mistake.}
    \item \texttt{For example, if Action 3 is the first wrong action, output \textless turn\_idx\textgreater 3\textless /turn\_idx\textgreater.}
    \item \texttt{The integer inside \textless turn\_idx\textgreater{} must be from 1 to \textcolor{varcolor}{\{max\_turn\}}, matching the action numbers shown above.}
\end{itemize}
\end{tcolorbox}
    \caption{The pivotal reflection prompt template for the Minesweeper agent.}
    \label{fig:minesweeper_pivotal_reflection_prompt}
\end{figure*}

\begin{figure*}[h]
    \centering
\begin{tcolorbox}[
    arc=1mm,
    auto outer arc,
    title={\textbf{ALFRED Play Prompt}},
    width=\textwidth,
    colback=white,
    colframe=black!75!white,
    boxrule=0.5mm
]
% --- Prompt Section ---
\noindent \texttt{You are an expert agent operating in the ALFRED Embodied Environment.}\\
\texttt{\textcolor{varcolor}{\{init\_observation\}}\textcolor{varcolor}{\{past\_trajectories\_reflections\}}\textcolor{varcolor}{\{trajectory\}}}

\vspace{0.5em}
\noindent \textbf{\# Admissible Actions}\\
\texttt{Your admissible actions of the current situation are:}\\
\texttt{[\textcolor{varcolor}{\{admissible\_actions\}}]}

\vspace{0.5em}
\noindent \textbf{\# Your Task}\\
\noindent \texttt{Now it's your turn to take an action.}

\vspace{0.5em}
\begin{itemize}[leftmargin=*, noitemsep, topsep=0pt]
    \item \texttt{If previous reflections are provided, use them as guidance but re-evaluate the current state and justify your choice with your own reasoning.}
    \item \texttt{Do not copy a prior reflection or past action sequence verbatim; adapt principles to the current context.}
    \item \texttt{Your response should first provide step-by-step reasoning about the current situation.}
    \item \texttt{Once you have finished your reasoning, choose an admissible action for the current step and present it within \textcolor{tagcolor}{<action>} \textcolor{tagcolor}{</action>} tags.}
\end{itemize}

\vspace{0.5em}
\noindent \textbf{Format:}\\
\texttt{your step-by-step reasoning}\\
\texttt{\textcolor{tagcolor}{<action>}admissible action\textcolor{tagcolor}{</action>}}

\end{tcolorbox}
    \caption{The play prompt template for the ALFRED embodied agent.}
    \label{fig:alfred_play_prompt}
\end{figure*}

\begin{figure*}[h]
    \centering
\begin{tcolorbox}[
    arc=1mm,
    auto outer arc,
    title={\textbf{ALFWorld Pivotal Reflection Prompt}},
    width=\textwidth,
    colback=white,
    colframe=black!75!white,
    boxrule=0.5mm
]
% --- Prompt Content ---
\noindent \texttt{Now reflect on the past experience:}

\begin{itemize}[leftmargin=*, noitemsep, topsep=0pt]
    \item \texttt{First, analyze each turn's decision and identify the \textbf{root cause} of failure.}
    \item \texttt{Then determine the \textbf{earliest action number} where the trajectory went wrong}
    \item \texttt{(the first shown action that should be changed).}
    \item \texttt{Finally, provide your reflection with the retry point.}
\end{itemize}

\noindent \texttt{Your response MUST end with:}
\begin{enumerate}[leftmargin=*, noitemsep, topsep=0pt]
    \item \texttt{A \textless turn\_idx\textgreater{} tag containing the earliest turn number to retry from.}
    \item \texttt{A \textless remark\textgreater{} tag containing your concise reflection and improved plan.}
\end{enumerate}

\noindent \texttt{Format:}\\
\texttt{\textless turn\_idx\textgreater integer\textless /turn\_idx\textgreater}\\
\texttt{\textless remark\textgreater your concise reflection and improved plan \textless /remark\textgreater}\\
\noindent \texttt{Where:}
\begin{itemize}[leftmargin=*, noitemsep, topsep=0pt]
    \item \texttt{The text inside \textless remark\textgreater{} is a concise improved plan that accounts for the identified mistake.}
    \item \texttt{For example, if Action 3 is the first wrong action, output \textless turn\_idx\textgreater 3\textless /turn\_idx\textgreater.}
    \item \texttt{The integer inside \textless turn\_idx\textgreater{} must be from 1 to \textcolor{varcolor}{\{max\_turn\}}, matching the action numbers shown above.}
\end{itemize}
\end{tcolorbox}
    \caption{The pivotal reflection prompt template for the ALFRED embodied agent.}
    \label{fig:alfworld_pivotal_reflection_prompt}
\end{figure*}

\begin{figure*}[h]
    \centering
\begin{tcolorbox}[
    arc=1mm,
    auto outer arc,
    title={\textbf{SearchQA Play Prompt}},
    width=\textwidth,
    colback=white,
    colframe=black!75!white,
    boxrule=0.5mm
]
% --- Prompt Section ---
\noindent \texttt{You are an expert agent tasked with answering the given question step-by-step.}\\
\texttt{Your question: \textcolor{varcolor}{\{task\_description\}}}

\vspace{0.5em}
\noindent \textbf{\# Your Task}\\
\noindent \texttt{Now it's your turn to respond for the current step.}\\
\texttt{You should first conduct reasoning process. This process MUST be enclosed within \textcolor{tagcolor}{<think>} \textcolor{tagcolor}{</think>} tags.}

\vspace{0.5em}
\noindent \textbf{\# Action Space}\\
\noindent \texttt{After completing your reasoning, choose only one of the following actions. Do not perform both:}
\begin{enumerate}[leftmargin=*, noitemsep, topsep=0pt]
    \item \texttt{If you lack some knowledge, call a search engine to get external information using the format: \textcolor{tagcolor}{<search>} your query \textcolor{tagcolor}{</search>}.}
    \item \texttt{If you have enough knowledge to answer confidently, provide your final answer within \textcolor{tagcolor}{<answer>} \textcolor{tagcolor}{</answer>} tags, without detailed illustrations.}
\end{enumerate}

\vspace{0.5em}
\noindent \textbf{\# Reflection}\\
\noindent \texttt{On trial \#\textcolor{varcolor}{\{traj\_idx\}}, the task was not solved well enough. Your reflection is:}\\
\texttt{\textcolor{varcolor}{\{reflection\}}}

\vspace{0.5em}
\noindent \textbf{Format:}\\
\texttt{\textcolor{tagcolor}{<think>}your reasoning process\textcolor{tagcolor}{</think>}}\\
\texttt{\textcolor{tagcolor}{<search>}your query\textcolor{tagcolor}{</search>}}\\
\texttt{or}\\
\texttt{\textcolor{tagcolor}{<answer>}final answer\textcolor{tagcolor}{</answer>}}

\end{tcolorbox}
    \caption{The play prompt template for the SearchQA agent.}
    \label{fig:searchqa_play_prompt}
\end{figure*}

\begin{figure*}[h]
    \centering
\begin{tcolorbox}[
    arc=1mm,
    auto outer arc,
    title={\textbf{SearchQA Pivotal Reflection Prompt}},
    width=\textwidth,
    colback=white,
    colframe=black!75!white,
    boxrule=0.5mm
]
% --- Prompt Section ---
\noindent \texttt{You are an expert agent tasked with answering the given question step-by-step.}\\
\texttt{Your question: \textcolor{varcolor}{\{task\_description\}}}

\noindent \textbf{\# Your Task}\\
\noindent \texttt{You will be given the history of a past attempt.}\\
\texttt{Your job is to:}
\begin{enumerate}[leftmargin=*, noitemsep, topsep=0pt]
    \item \texttt{Reflect on the past search process and identify the earliest turn where a wrong or inefficient decision was made.}
    \item \texttt{Determine the earliest turn number whose action should be changed.}
    \item \texttt{Devise a concise improved plan starting from that turn.}
\end{enumerate}

\noindent \textbf{\# Past Experience}\\
\texttt{\textcolor{varcolor}{\{current\_trajectory\}}}\\
\texttt{The task is NOT successfully completed.}

\noindent \texttt{Now reflect on the past experience:}

\begin{itemize}[leftmargin=*, noitemsep, topsep=0pt]
    \item \texttt{First, analyze the search process turn by turn and identify the root cause of failure.}
    \item \texttt{Then determine the earliest action number where the trajectory went wrong.}
    \item \texttt{The earliest action is the first shown search or answer step that should be changed.}
    \item \texttt{Focus on the earliest pivotal mistake, not a later downstream consequence.}
    \item \texttt{Finally, provide a concise improved plan for retrying from that turn.}
\end{itemize}

\noindent \textbf{Format:}\\
\texttt{\textcolor{tagcolor}{<turn\_idx>}integer\textcolor{tagcolor}{</turn\_idx>}}\\
\texttt{\textcolor{tagcolor}{<remark>}}\\
\texttt{your concise reflection and improved plan}\\
\texttt{\textcolor{tagcolor}{</remark>}}

\noindent \textbf{Where:}
\begin{itemize}[leftmargin=*, noitemsep, topsep=0pt]
    \item \texttt{The text inside \textcolor{tagcolor}{<remark>} should summarize the corrected search strategy, missing checks, or improved answer criterion for the retry.}
    \item \texttt{The integer inside \textcolor{tagcolor}{<turn\_idx>} must be from 1 to \textcolor{varcolor}{\{max\_turn\}}.}
    \item \texttt{Use 1 if the very first shown step should be changed.}
    \item \texttt{For example, if Step 3 is the first wrong search or answer decision, output \textcolor{tagcolor}{<turn\_idx>}3\textcolor{tagcolor}{</turn\_idx>}.}
\end{itemize}

\end{tcolorbox}
    \caption{The pivotal reflect prompt template for the SearchQA agent.}
    \label{fig:searchqa_reflection_prompt}
\end{figure*}

\begin{figure}[t] % [t] 表示放置在页顶，也可以用 [b] 放页底
    \centering % 可选：居中（虽然 tcolorbox 设置了全宽，但加上是个好习惯）
\begin{tcolorbox}[
    colback=white,
    colframe=darkblue,
    width=\textwidth,
    sharp corners,
    boxrule=0.8pt,
    title=\textbf{Case: Minesweeper (Pivotal Retry)},
    fonttitle=\sffamily\bfseries,
    colbacktitle=lightgray,
    coltitle=black,
    left=5pt, right=5pt, top=5pt, bottom=5pt
]
\small

\textbf{Task:} Reveal safe cells and avoid hidden mines (6$\times$6 board, 3 mines).

\begin{tcolorbox}[colback=lightgray!50, colframe=lightgray!50, boxrule=0pt, left=2pt, right=2pt, top=2pt, bottom=2pt]
    \centering\textbf{\textsf{Failed Agent Trajectory in First Attempt}}
\end{tcolorbox}

\begin{tabularx}{\linewidth}{p{1.4cm} X}
    \rowcolor{clean-green!10} \textbf{Step 1} & \textbf{Obs:}  \\
    & \texttt{Row 1: 2 ? ? ? ? ?}\\
    & \texttt{Row 2: ? ? ? ? ? ?}\\
    & \texttt{Row 3: ? 1 ? ? ? ?}\\
    & \texttt{Row 4: ? ? ? ? ? ?}\\
    & \texttt{Row 5: ? ? ? ? ? ?}\\
    & \texttt{Row 6: ? ? ? ? ? ?}\\
    & \textbf{Think:} Step-by-Step Reasoning: We are given a 6x6 Minesweeper board with 3 hidden mines. The only known cell is at (3,2), which is a numbered cell with the value '1' ... \\
    & \textbf{Action:} \texttt{(3, 3)} \\
    \midrule

    \rowcolor{clean-green!10} \textbf{Step 2} & \textbf{Obs:} \\
    & \texttt{Row 1: . . 1 ? 1 .}\\
    & \texttt{Row 2:  . . 1 1 1 .}\\
    & \texttt{Row 3: 1 1 . . . .}\\
    & \texttt{Row 4: ? 1 . . . .}\\
    & \texttt{Row 5: ? 2 . . . .}\\
    & \texttt{Row 6: ? 1 . . . .}\\
    & \textbf{Think:} Step-by-Step Reasoning: Let's analyze the current state of the board after the first action. We need to deduce where the remaining **3 ... \\
    & \textbf{Action:} \texttt{(3, 2)} \\
    \midrule

    \rowcolor{clean-green!10} \textbf{Step 3} & \textbf{Obs:} \\
    & \texttt{Row 1: . . 1 ? 1 .}\\
    & \texttt{Row 2:  . . 1 1 1 .}\\
    & \texttt{Row 3: 1 1 . . . .}\\
    & \texttt{Row 4: ? 1 . . . .}\\
    & \texttt{Row 5: ? 2 . . . .}\\
    & \texttt{Row 6: ? 1 . . . .}\\
    & \textbf{Think:} Step-by-Step Reasoning: Let's analyze the current state of the board after the first action. We need to deduce where the remaining **3 ... \\
    & \textbf{Action:} \texttt{(4, 1)} \\

    \vspace{2pt}
    \textbf{\textcolor{heat-red}{Failed}} & \vspace{2pt}\textcolor{heat-red}{First attempt fails after an unsafe pivot decision branch.}
\end{tabularx}

\begin{tcolorbox}[colback=lightgray!50, colframe=lightgray!50, boxrule=0pt, left=2pt, right=2pt, top=2pt, bottom=2pt]
    \centering\textbf{\textsf{Self-Reflect on Failed Trajectory}}
\end{tcolorbox}

\begin{tabularx}{\linewidth}{p{1.4cm} X}
    \rowcolor{darkblue!10} \textbf{Reflection} & \textbf{Reasoning:} After Action 1 clicking (3,3), the board reveals substantial information. Examining the neighbors of (3,1)=1: among its 8 surrounding cells, (2,1), (2,2), and (3,2) are all revealed ... \\
    \rowcolor{darkblue!10} & \textbf{Retry Turn:} \texttt{<turn\_idx>3</turn\_idx>}  \\
    \rowcolor{darkblue!10} & \textbf{Experience:} The mistake was revealing (4, 1) which turned out to be a mine. The improved plan is to avoid guessing based on insufficient information and instead look for cells with unique numbers that can be deduced safely. 
\end{tabularx}

\begin{tcolorbox}[colback=lightgray!50, colframe=lightgray!50, boxrule=0pt, left=2pt, right=2pt, top=2pt, bottom=2pt]
    \centering\textbf{\textsf{Success Agent Trajectory in Reflect Retry}}
\end{tcolorbox}

\begin{tabularx}{\linewidth}{p{1.4cm} X}
    \rowcolor{clean-green!10} \textbf{Step 3} & \textbf{Obs:}   \\
    & \texttt{Row 1: . . 1 ? 1 .}\\
    & \texttt{Row 2: . . 1 1 1 .}\\
    & \texttt{Row 3: 1 1 . . . .}\\
    & \texttt{Row 4: ? 1 . . . .}\\
    & \texttt{Row 5: ? 2 . . . .}\\
    & \texttt{Row 6: ? 1 . . . .}\\
    & \textbf{Think:} <reflection>The previous attempt revealed a mine at (4, 1), which was a mistake. This highlights the need to be cautious and rely on logical deduction rather than random guessing.Looking at the current board state, I notice that some cells have been revealed and are numbered. \\
    & \textbf{Action:} \texttt{(5, 1)} \\
    \vspace{2pt}
    \textbf{\textcolor{clean-green}{Success}} & \vspace{2pt}\textcolor{clean-green}{Reflect retry succeeds from the same pivot observation state.}
\end{tabularx}
\end{tcolorbox}
    \caption{Case: Minesweeper (Pivotal Retry)}
    \label{app:case_minesweeper}
\end{figure}

\begin{figure*}[t] % [t] 表示放置在页顶，也可以用 [b] 放页底
    \centering % 可选：居中（虽然 tcolorbox 设置了全宽，但加上是个好习惯）
% --- ALFWorld Case 1: Clean Task ---
\begin{tcolorbox}[
    colback=white,
    colframe=darkblue,
    width=\textwidth,
    sharp corners,
    boxrule=0.8pt,
    title=\textbf{Case: ALFWorld SprayBottle (Pivotal Retry)},
    fonttitle=\sffamily\bfseries,
    colbacktitle=lightgray,
    coltitle=black,
    left=5pt, right=5pt, top=5pt, bottom=5pt
]
\small

\textbf{Task:} put two spraybottle in cabinet.

\begin{tcolorbox}[colback=lightgray!50, colframe=lightgray!50, boxrule=0pt, left=2pt, right=2pt, top=2pt, bottom=2pt]
    \centering\textbf{\textsf{Failed Agent Trajectory in First Attempt}}
\end{tcolorbox}

\begin{tabularx}{\linewidth}{p{1.4cm} X}
    \rowcolor{clean-green!10} \textbf{Step 1} & \textbf{Obs:} Room initialized; countertop 1 contains spraybottle 1 and spraybottle 2. \\
    & \textbf{Think:} To get two spray bottles, I need to check possible places where they could be located. Let me start checking possible locations where they might be inside. \\
    & \textbf{Action:} \texttt{go to countertop 1} \\
    \midrule

    \rowcolor{clean-green!10} \textbf{Step 2} & \textbf{Obs:} spraybottle 1 is available on countertop 1. \\
    & \textbf{Think:} To put two spraybottle in cabinet, I need to take two spraybottle. On countertop 1, I notice spraybottle 1 and spraybottle 2 are available. \\
    & \textbf{Action:} \texttt{take spraybottle 1 from countertop 1} \\
    \midrule

    \rowcolor{clean-green!10} \textbf{Step 3} & \textbf{Obs:} Agent holds spraybottle 1 and reaches cabinet area. \\
    & \textbf{Think:} I now hold spraybottle 1. To put it in a cabinet, I can choose any of the available cabinets. \\
    & \textbf{Action:} go to cabinet 1 \\
    \midrule

    \rowcolor{clean-green!10} \textbf{Step 4} & \textbf{Obs:}  You open the cabinet 1. The cabinet 1 is open. In it, you see nothing\\
    & \textbf{Think:} I currently hold spraybottle 1. The next step is to put it inside a cabinet. \\
    & \textbf{Action:} open cabinet 1 \\
    \midrule

    \rowcolor{heat-red!10} \textbf{Step 5} & \textbf{Obs:} \textit{Nothing happens.} \\
    & \textbf{Think:} I currently am at cabinet 1. I need to place two spraybottle in it.  \\
    & \textbf{Action:} take spraybottle 1 from countertop 1 \\
    \vspace{2pt}
    \textbf{\textcolor{heat-red}{Result: Failed}} & \vspace{2pt}\textcolor{heat-red}{The first attempt fails after ineffective action at the pivot state.}
\end{tabularx}

\begin{tcolorbox}[colback=lightgray!50, colframe=lightgray!50, boxrule=0pt, left=2pt, right=2pt, top=2pt, bottom=2pt]
    \centering\textbf{\textsf{Self-Reflect on Failed Trajectory}}
\end{tcolorbox}

\begin{tabularx}{\linewidth}{p{1.4cm} X}
    \rowcolor{darkblue!10} \textbf{Reflection} & \textbf{Reasoning:} Looking at the history, I notice that the turn where the mistake was first made is Turn 5. On that turn, I attempted to place spraybottle 1 into cabinet 1, but it didn't go into the cabinet. Additionally, I note that there are still items on countertops where items should not be placed. \\
    \rowcolor{darkblue!10} & \textbf{Retry Turn:} Step 4 \\
    \rowcolor{darkblue!10} & \textbf{Experience:} Use corrected placement action, then fetch and place the second spraybottle.
\end{tabularx}

\begin{tcolorbox}[colback=lightgray!50, colframe=lightgray!50, boxrule=0pt, left=2pt, right=2pt, top=2pt, bottom=2pt]
    \centering\textbf{\textsf{Success Agent Trajectory in Reflect Retry}}
\end{tcolorbox}

\begin{tabularx}{\linewidth}{p{1.4cm} X}
    \rowcolor{clean-green!10} \textbf{Step 4} & \textbf{Obs:} You open the cabinet 1. The cabinet 1 is open. In it, you see nothing. \\
    & \textbf{Think:} I will do that step by step. Looking around, I notice that \"put spraybottle 1 in/on cabinet 1\" is available. \\
    & \textbf{Action:} \texttt{put spraybottle 1 in/on cabinet 1} \\
    \midrule

    \rowcolor{clean-green!10} \textbf{Step 5} & \textbf{Obs:} You put the spraybottle 1 in/on the cabinet 1. \\
    & \textbf{Think:} I currently hold spraybottle 1. \\
    & \textbf{Action:} go to countertop 1\\
    \midrule

    \rowcolor{clean-green!10} \textbf{Step 6} & \textbf{Obs:} Holding spraybottle 2. \\
    & \textbf{Think:} - Current observations show I need to put two spraybottle in a cabinet. So far, spraybottle 1 has been placed in cabinet 1. To complete the task, there is another spraybottle to place in a cabinet. \\
    & \textbf{Action:} take spraybottle 2 from countertop 1 \\
    \midrule

    \rowcolor{clean-green!10} \textbf{Step 7} & \textbf{Obs:} You pick up the spraybottle 2 from the countertop 1. \\
    & \textbf{Think:} - I currently hold spraybottle 2. Now, to place it in a cabinet, I can do so to an available location.  \\
    & \textbf{Action:} go to cabinet 1 \\
    \midrule

    \rowcolor{clean-green!10} \textbf{Step 8} & \textbf{Obs:} You arrive at loc 10. The cabinet 1 is open. In it, you see a spraybottle 1. \\
    & \textbf{Think:} - I currently hold the spray bottle called spraybottle 2. \\
    & \textbf{Action:} put spraybottle 2 in/on cabinet 1 \\

    \vspace{2pt}
    \textbf{\textcolor{clean-green}{Result: Success}} & \vspace{2pt}\textcolor{clean-green}{Reflect retry succeeds from the same pivot observation state.}
\end{tabularx}
\end{tcolorbox}
    \caption{Case: ALFWorld Example (Pivotal Retry)}
    \label{app:case_webshop}
\end{figure*}

\begin{figure*}[t]
    \centering
\begin{tcolorbox}[
    colback=white,
    colframe=darkblue,
    width=\textwidth,
    sharp corners,
    boxrule=0.8pt,
    title=\textbf{Case: Complex Search QA (Pivotal Retry)},
    fonttitle=\sffamily\bfseries,
    colbacktitle=lightgray,
    coltitle=black,
    left=5pt, right=5pt, top=5pt, bottom=5pt
]
\small

\textbf{Task:} How tall is the ride that is nearly identical to the Goliath at Six Flags Over Texas?

\begin{tcolorbox}[colback=lightgray!50, colframe=lightgray!50, boxrule=0pt, left=2pt, right=2pt, top=2pt, bottom=2pt]
    \centering\textbf{\textsf{Failed Agent Trajectory in First Attempt}}
\end{tcolorbox}

\begin{tabularx}{\linewidth}{p{1.3cm} X}
    \rowcolor{clean-green!10} \textbf{Step 1} & \textbf{Obs:} Goliath at Six Flags Fiesta Texas (formerly Batman: The Ride) is an inverted roller coaster originally from Japan. Meanwhile, Goliath at Six Flags Magic Mountain is described as a steel hypercoaster ``nearly identical to Titan at Six Flags Over Texas,'' though its height is not given in the returned snippet. \\
    & \textbf{Think:} The search results indicate that Goliath at Six Flags Magic Mountain is nearly identical to Titan at Over Texas, but the height of Goliath at Magic Mountain is not provided. I need to look up the specific height of that ride. \\
    & \textbf{Action:} \texttt{search: How tall is Goliath at Six Flags Magic Mountain?} \\
    \midrule

    \rowcolor{clean-green!10} \textbf{Step 2} & \textbf{Obs:} The result describes Goliath at Magic Mountain as a hypercoaster with a sub-tropical Mayan theme. It states ``Goliath lacks the 540-degree upward helix on Titan prior to the mid-course brake run,'' again confirming the nearly-identical relationship to Titan. However, the search snippet still does not enumerate a numeric height value. \\
    & \textbf{Think:} The information confirms Titan at Over Texas is Goliath at Magic Mountain's twin. The height of Goliath at Magic Mountain is not explicitly mentioned in the returned documents, so I will rely on my prior knowledge. \\
    & \textbf{Action:} \texttt{answer: 175 feet} \\
    \vspace{2pt}
    \textbf{\textcolor{heat-red}{Result: Failed}} & \vspace{2pt}\textcolor{heat-red}{The agent conjectures a height without verification and answers incorrectly---the true target is Titan, not Goliath at Magic Mountain.}
\end{tabularx}

\begin{tcolorbox}[colback=lightgray!50, colframe=lightgray!50, boxrule=0pt, left=2pt, right=2pt, top=2pt, bottom=2pt]
    \centering\textbf{\textsf{Self-Reflect on Failed Trajectory}}
\end{tcolorbox}

\begin{tabularx}{\linewidth}{p{1.3cm} X}
    \rowcolor{darkblue!10} \textbf{Reflection} & \textbf{Reasoning:} The initial search (Step~1) revealed that Goliath at Magic Mountain is nearly identical to Titan at Over Texas. The pivotal mistake occurred at Step~2, where the agent searched for the height of Goliath at Magic Mountain instead of directly searching for Titan. The search for Goliath at Magic Mountain only returned a feature description without a height value, yet the agent still produced an unverified answer of ``175 feet.'' The correct strategy is to search for Titan's height, since Titan---not Goliath at Magic Mountain---is the ride that the question implicitly asks about. \\
    \rowcolor{darkblue!10} & \textbf{Pivot Turn:} Step~2 \\
    \rowcolor{darkblue!10} & \textbf{Improved Plan:} Retain the Step~1 observation (the TItan--Goliath relationship), then replace Step~2 with a direct search for ``height of Titan at Six Flags Over Texas.''
\end{tabularx}

\begin{tcolorbox}[colback=lightgray!50, colframe=lightgray!50, boxrule=0pt, left=2pt, right=2pt, top=2pt, bottom=2pt]
    \centering\textbf{\textsf{Success Agent Trajectory in Reflect Retry (Retry from Step~2)}}
\end{tcolorbox}

\begin{tabularx}{\linewidth}{p{1.3cm} X}
    \rowcolor{darkblue!10} \textbf{Step 2} & \textbf{Obs:} Titan is a steel hyper coaster located at Six Flags Over Texas in Arlington. As described in the search result, ``It stands at 245 feet and contains a 255-foot drop at 85 miles per hour. It is the tallest, fastest, and longest coaster in Texas.'' \\
    \textbf{\textcolor{heat-red}{(corrected)}} & \textbf{Think:} The ride nearly identical to Goliath at Over Texas is Titan at Over Texas. The search directly returns Titan's height as 245 feet. \\
    & \textbf{Action:} \texttt{search: height of Titan at Six Flags Over Texas} \\
    \midrule

    \rowcolor{clean-green!10} \textbf{Step 3} & \textbf{Obs:} \textit{The search result clearly states Titan stands at 245 feet.} \\
    & \textbf{Think:} Titan, the ride nearly identical to Goliath at Over Texas, is confirmed to be 245 feet tall. This directly answers the question. \\
    & \textbf{Action:} \texttt{answer: 245 feet} \\
    \vspace{2pt}
    \textbf{\textcolor{clean-green}{Result: Success}} & \vspace{2pt}\textcolor{clean-green}{Retrying from the pivot at Step~2 with the corrected search target yields the correct answer.}
\end{tabularx}
\end{tcolorbox}
    \caption{Case: SearchQA Example (Pivotal Retry)}
    \label{app:case_search}
\end{figure*}

\end{document}